\documentclass[12pt,letterpaper]{article}
\usepackage[a4paper, total={7in, 10in}]{geometry}

\usepackage{graphicx}
\usepackage{helvet}
\usepackage{authblk}
\usepackage{hyperref}
\usepackage{amsmath} 
\usepackage{amssymb} 
\usepackage{orcidlink} 
\usepackage[super,comma,sort&compress]  
   {natbib}
\usepackage[right]{lineno} 
\usepackage{pdfpages}
\usepackage{glossaries}
\usepackage{textcomp}
\usepackage[T1]{fontenc}

\newacronym{ar}{AR}{augmented reality}
\newacronym{mr}{MR}{mixed reality}
\newacronym{vr}{VR}{virtual reality}
\newacronym{tlx}{TLX}{Task Load Index}
\newacronym{sus}{SUS}{System Usability Scale}

\makeatletter
\renewcommand{\maketitle}{\bgroup\setlength{\parindent}{0pt}
\begin{flushleft}
  \textbf{\@title}
  
  \@author
\end{flushleft}\egroup}
\makeatother

\title{A Wearable Pneumatic Device for Continuous, Closed-Loop, Bidirectional Tactile Interaction}
\date{}

\author[1,+,*\orcidlink{0000-0000-0000-0000}]{Cosima du Pasquier}
\author[1,+\orcidlink{0009-0009-7758-1097}]{Aliyah Smith}
\author[1]{Serin Huber}
\author[1]{Joshua Phelps}
\author[1]{Ilana A. Cohen}
\author[1,\orcidlink{0000-0001-5972-5311}]{Ava Chen}
\author[1]{Monroe Kennedy, III}
\author[1,\orcidlink{0000-0002-6912-1666}]{Allison M. Okamura}

\affil[1]{Department of Mechanical Engineering, Stanford University, Stanford, CA, USA}
\affil[+]{These authors contributed equally}

\affil[*]{Correspondence: cosimad@stanford.edu}

\begin{document}

\maketitle

\section*{SUMMARY}

We present a system of two wearable pneumatic haptic devices that supports continuous, closed-loop, bidirectional tactile interaction at perceptually relevant force and temporal scales. A single device can contain up to twelve pressure sensing channels connected to textile-based pneumatic pouches. Each channel in a device can be used as a sensor, an actuator, or both. As an actuator with integrated sensing, the channel generates stable skin indentation through local closed-loop control. As a sensor, a channel can be mounted (or worn) on any surface, including on a robot gripper or on the human body, and used to measure touch interactions with the environment or a human user. A distributed architecture supports sustained pressure output, rapid dynamic response, and wireless pairing of identical devices in a system to transmit and reproduce tactile pressure signals in real time.

Device-level characterization demonstrates force bandwidth exceeding 30 Hz, rapid and well-damped step responses, and extended pressure retention compared to prior compact pneumatic platforms. Human studies show that pressure-based fingertip feedback enables discrimination of force and stiffness, improves teleoperated manipulation by reducing applied pressures by up to 23.1\% and task duration by up to 27.4\%, and lowers subjective mental workload by 18.8\%, particularly under visually constrained conditions. By unifying tactile sensing and haptic feedback within a single pneumatic modality, the device provides a practical foundation for bidirectional touch interaction in teleoperation.

\section*{KEYWORDS}


Soft Robotics, Haptics, Teleoperation

\section*{INTRODUCTION}

\begin{figure}[h]
    \centering
    \includegraphics[width=0.9\linewidth]{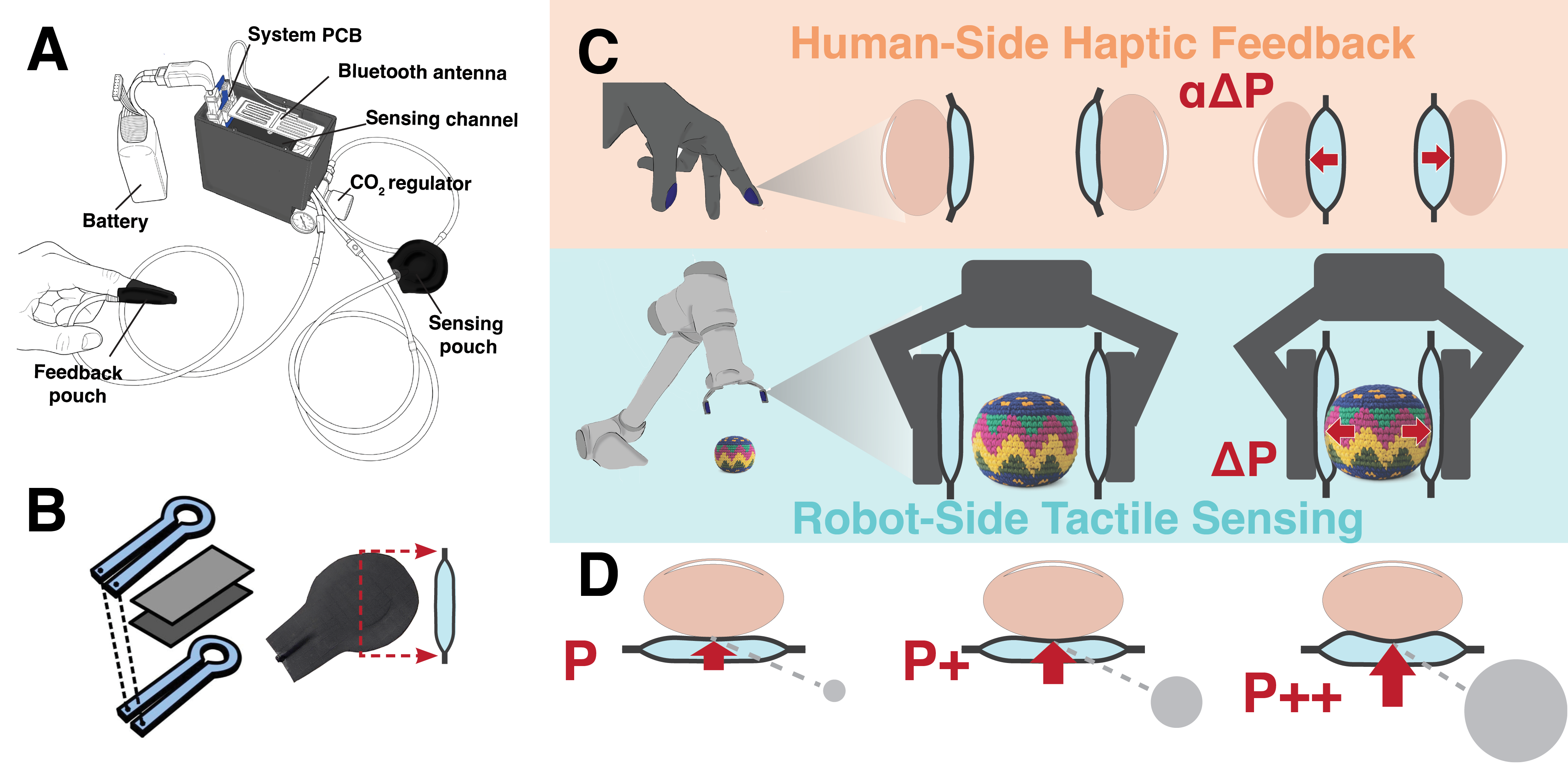}
    \caption{\textbf{A} Full untethered pneumatic device overview with two channels: one sensing pouch for the robot, and one haptic feedback pouch worn by a user. \textbf{B} The pouches are made by die casting TPU coated nylon. \textbf{C} Two devices are connected in a twin configuration to enable haptic feedback during teleoperation. The user's hand motion is tracked and sent as a target configuration for the pose and grasp for a robot manipulator. The pouches on the robot side sense interaction with an object (here, a hackey~sack), and the pressure changes are forwarded with gain $\alpha$ to pouches worn by the user. \textbf{D} When the pouch is inflated while pressing on the fingertip, the user feels both increasing normal pressure and an increase in contact area.}
    \label{fig:overview}
\end{figure}

The human sense of touch is inherently bidirectional: we respond physically to what we feel in real time. Applying this two-way exchange to teleoperation improves performance on several levels: haptic feedback lowers task completion time~\cite{Pacchierotti2012TwoFinger,Cuan2024}, improves manipulation accuracy~\cite{Okamura2004}, and decreases both peak and average forces applied during interaction~\cite{Wagner2002}. Yet in practice the two directions are engineered separately: haptic displays are designed for humans and tactile sensors are developed for robots, and the mismatch between what the robot records and what the operator feels limits both~\cite{Cuan2024}. Here, we present a wearable pneumatic device in which sensing and display use the same mechanism: a textile pouch whose internal pressure is simultaneously the sensed and the actuated variable. Paired wirelessly, two such devices reproduce in real time the pressure sensed at each gripper finger on the corresponding fingertip of the robot operator (Fig.~\ref{fig:overview}).

As teleoperation is increasingly used not only for control but also as a source of demonstrations for imitation learning, continuous, interpretable, bidirectional tactile feedback has become essential. To understand why it has remained rare requires examining both ends of the connection: how touch is displayed at the hand, and how it is sensed at the robot.

On the operator, haptic feedback is conventionally divided into two types: kinesthetic and cutaneous. Kinesthetic feedback conveys force to the muscles and joints, typically through world-grounded interfaces. For teleoperation, this relies on bilateral position-exchange control or on expensive, fragile force sensors~\cite{Okamura2004}, and bilateral control loops are destabilized by hard contacts, stiff settings, and communication latency~\cite{Lawrence1993,HashtrudiZaad2002}; stability is recovered only by scaling down force, which sacrifices the transparency ideally provided by kinesthetic feedback~\cite{Pacchierotti2014TAP}. In contrast, cutaneous feedback is delivered directly to the skin and underlying tissues. Cutaneous stimulation at the fingerpad can be ungrounded, such that forces are reacted by the user's own finger. Ungrounded, wearable cutaneous feedback for teleoperation has two advantages: the device is wearable and free of workspace constraints, and keeps the teleoperation loop intrinsically stable even under hard contacts or communication delay~\cite{Meli2014,Pacchierotti2023}. How that cutaneous information is delivered then determines its interpretability. The dominant approach, \textit{sensory substitution}, encodes the missing force into an unrelated channel such as abstract vibration, an auditory beep, or brightness~\cite{ErkatBatuhan2025}. Because the cue bears no natural relationship to the interaction, the human operator must learn the mapping, and such cues poorly encode continuous quantities like force magnitude or compliance~\cite{Cui2025,Yu2022}, increasing cognitive load~\cite{Ferguson2021}. Moreover, when teleoperation with sensory substitution is used to collect data for imitation and diffusion learning, the training data is noisy and degrades learned policies~\cite{Cuan2024,Hu2025}. For the low-force interactions characteristic of dexterous manipulation, local skin deformation feedback is better suited because it provides cutaneous stimulation akin to real-world contact~\cite{Prattichizzo2013,Quek2015}, rather than a cue the operator must decode.

At the robot, tactile sensing measures the state of contact, potentially including contact geometry, force, texture, and slip~\cite{Di2024}, most commonly through vision-based sensors that image the deformation of an illuminated elastomer~\cite{Yuan2017,Di2025,Zhao2025} or through dense electronic arrays~\cite{Nie2026,Dahiya2010,Chang2025,Uzunoglu2026}. The defining tension is a reciprocal one between spatial resolution and speed: denser sensing captures finer contact geometry but produces high-bandwidth data streams that impose substantial sensing, transmission, and processing overhead, while faster sampling can be achieved only by coarsening spatial resolution~\cite{Lin2023,Yao2025,Zhu2025}. In practice, the tactile stream is distilled during teleoperation to a single contact scalar, typically normal force amplitude or incipient slip~\cite{Lippi2024,Mukashev2025,Zhang2025,Liang2026}, and mapped to fingertip vibration, or preserved spatially and mapped to a tactile array~\cite{Jia2026}, vibration across multiple fingers~\cite{Xu2025}, or directional rotation of a vibration motor~\cite{Tan2026}. In every case, the robot records one signal modality and the operator feels another.

Pneumatics is unusual in being able to serve both ends at once. The same inflatable textile pouch that displays normal force through direct skin indentation~\cite{ErkatBatuhan2025,duPasquier2024,Jumet2023} can measure it, because the pressure inside the pouch is simultaneously the actuation variable and the sensed one (Fig.~\ref{fig:overview}A and B). Beyond conveying normal force, an inflated contact reproduces the natural growth of contact area with force that the fingerpad experiences against a real compliant object (Fig.~\ref{fig:overview}D), long understood as a primary cue for softness discrimination~\cite{Bicchi2000}. What has stood between pneumatics and this promise is control: existing pneumatic controllers are bulky, due to external compressors and valves~\cite{Shtarbanov2021FlowIO,Jung2024UntetheredRobotics}, confining them to open-loop control in untethered settings~\cite{Sonar2020,Yu2022,Cianchetti2018}, while the compressibility of air makes stable, precise control difficult without feedback~\cite{Tang2026}. This absence of closed-loop control is a significant hurdle, since stable, sustained feedback is essential both for realistic perception and for reliable data collection.

Our device overcomes this hurdle with a compact, closed-loop controller that removes the compressor and valve burden that has historically tethered pneumatic devices, regulating each pouch locally so that pressure cues remain stable over extended interaction. In the twin architecture introduced above, sensed pressure is transmitted wirelessly with latency below the threshold at which tactile feedback is perceived as immediate, without intermediate modality change. We validate the design through device-level characterization and user studies, demonstrating reliable discrimination of transmitted force, improved teleoperation performance, and reduced subjective workload, establishing a practical foundation for bidirectional tactile communication that supports both human-in-the-loop control and data collection for learning-based robotic systems.

\section*{RESULTS}
The device is a wearable, untethered pneumatic system in which the same textile pouch can serve as a sensor or an actuator (Fig.~\ref{fig:overview}A and B). Three choices distinguish it from prior compact pneumatic haptic devices. Each pouch is paired with a proportional piezoelectric valve, a pressure sensor, and a dedicated microcontroller, so pressure is regulated in a local closed loop rather than by centralized open-loop solenoid switching (Table S2). The valve and air supply are integrated on the body of the device, removing the compressor and tether that have confined closed-loop pneumatic haptics to benchtop use. And because sensing and display share one physical modality, two devices can be paired so that pressure measured on one is reproduced on the other with no intermediate encoding other than amplification (Fig.~\ref{fig:overview}C).

We evaluate these choices at three levels. At the device level, the device holds regulated pressure for 30 min without a supply, settles to a new setpoint in 39 ms, delivers a -3 dB force response to 34.2 Hz and a peak force of 76.2 N, and mirrors pressure between paired devices in 64 ms. At the perceptual level, participants discriminated force and stiffness through the device as accurately as by hand, while weight discrimination degraded once three levels had to be ordered under teleoperated control. In teleoperated pick-and-place, feedback reduced applied pressure by up to 23.1\%, task duration by up to 27.4\%, and mental demand by 18.1\%, with the largest gains under visual occlusion.

Device airtightness was evaluated through pressure retention tests in which the external pressure supply was disconnected after full inflation. Across n = 5 trials, the device retained pressure for extended durations, with internal pressure decreasing from 2650 mbar to 2300 mbar after 5 min and remaining above 1700 mbar after 16 min before fully depressurizing after 30 min. In contrast, comparable compact pneumatic systems exhibited rapid pressure loss within seconds under identical conditions.

\subsection*{Operating at perceptually relevant force and temporal scales}
To assess whether the device operates at force and temporal scales relevant for tactile perception, we characterize its flow rate, force bandwidth, and maximum output force. These metrics determine the device’s ability to generate perceptually meaningful, time-varying normal forces at the fingertip.

\subsubsection*{Flow rate}
Inlet and outlet flow rates were estimated by recording the pressure change in a fixed volume chamber while running the device with $P_{inlet} = 1.4$ bar and venting to $P_{outlet} = -0.8$ bar, with the inlet and outlet valves respectively fully open, until the pressure reached a steady state. Wall air and vacuum were used in this experiment, though alternative air sources would yield similar results for the given inlet and outlet pressures. The detailed experimental procedure is provided in the Supplemental Information. A schematic of the system architecture is shown in Fig.~\ref{fig:systemarchitecture}.

The average inlet flow rate was 10.8 L/min, and the average outlet flow rate was 13.03 L/min, four to five times higher than other reported pneumatic controller flow rates (see Table S2). Given that the internal volume of the textile sensors and actuators used in this work is approximately $4 \cdot 10^{-3}$ L, these flow rates support rapid pressurization and depressurization, enabling high-frequency closed-loop control.

 \subsubsection*{Bandwidth}
The force bandwidth of the textile actuator was characterized by measuring its frequency-depen\-dent force response under blocked conditions. For n = 3 actuators, the device maintained a -3 dB force response up to 34.2 $\pm$ 1.2 Hz. This bandwidth exceeds that of many other untethered pneumatic haptic systems, which exhibit force roll-off below 15 Hz \cite{duPasquier2024,Delazio2018} (see SI Table S2); for a broader survey of soft actuation for wearable haptics, see \cite{Meng2026,Patel2025}.

From a perceptual standpoint, this bandwidth spans the range relevant for conveying dynamic tactile cues such as pulsing and force modulation, especially in the low-frequency regimes where conventional vibration-based approaches struggle to deliver strong forces. Human sensitivity to normal indentation and low-frequency force changes primarily involves the Meissner corpuscles and ranges from approximately 5~Hz to 50~Hz \cite{Piccinin2023,Culbertson2018}, making the measured response suitable for pressure-based tactile feedback.

\subsubsection*{Force}
Maximum actuator force was evaluated under blocked conditions at an input pressure of 2650 mbar. Across tested actuators, the device produced an average peak force of 76.2 ± 2.8 N. This force range exceeds the minimum thresholds required for reliable normal-force perception at the fingertip and lies within the range commonly used to study indentation-based tactile perception \cite{Ma2024,duPasquier2024}.

Because force output is regulated via closed-loop pressure control, the device enables continuous modulation of contact force rather than binary actuation. Prior work in haptics has shown that cutaneous force and skin deformation feedback can convey graded magnitude information such as stiffness and friction, whereas vibrotactile feedback is most commonly used to encode events or temporal patterns and has limited ability to represent force magnitude \cite{Culbertson2018}. Prior pneumatic haptic systems have predominantly relied on open-loop control, which limits feedback to transient pressure or force pulses rather than the sustained and continuously regulated contact forces required to replicate touch feedback during teleoperation \cite{duPasquier2024,Jumet2023}.

\subsection*{System responsiveness for bidirectional tactile communication}

\begin{figure}
    \centering
    \includegraphics[width=0.5\linewidth]{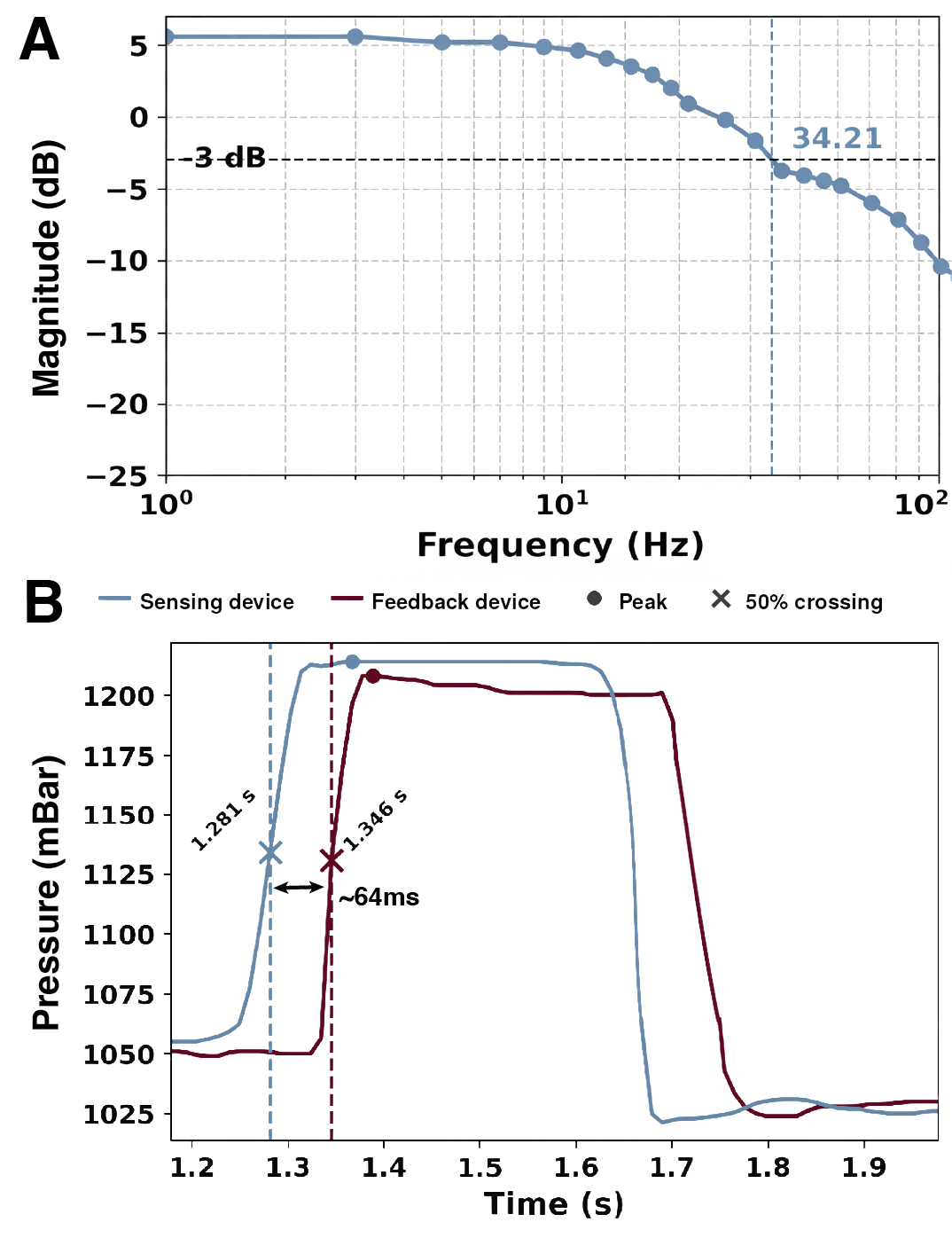}
    \caption{\textbf{Pneumatic Controller Characterization} Average \textbf{A} bandwidth and \textbf{B} transmission delay results in between two devices - one sensing device communicating to a haptic feedback device - for $n=10$ trials each.}
    \label{fig:characterization}
\end{figure}

To evaluate the suitability of the system for bidirectional tactile communication, we characterized both its local pressure response dynamics and its end-to-end latency.

System responsiveness was first assessed by measuring the intrinsic step response of the controller, isolating local actuation dynamics from communication effects. When subjected to a step input from a 1050 mbar baseline to 2000 mbar, the controller exhibited a rapid and well damped response, with an average 10–90\% rise time of 29 ms and a settling time of 39 ms (see Fig. S4).

End to end latency was then evaluated between two wirelessly connected devices by measuring the delay between a pressure event sensed on one device and the corresponding actuation response on a second device. Across repeated trials ($n = 10$), the average sensing to actuation latency was 64 ms (Fig.~\ref{fig:characterization}), with the dominant contribution arising from wireless communication via the host computer.

Taken together, these measurements place overall system responsiveness well below commonly cited perceptual thresholds for tactile feedback to be perceived as immediate (80–100 ms) \cite{immersion2013latency}. This performance supports real time bilateral and cooperative haptic interactions between distributed devices, including mirrored feedback, shared tactile experiences, and teleoperation scenarios requiring tight temporal coupling.

\subsection*{Haptic-enabled discrimination of force, stiffness, and weight in manual and mediated conditions.}
\begin{figure}
    \centering
    \includegraphics[width=0.9\linewidth]{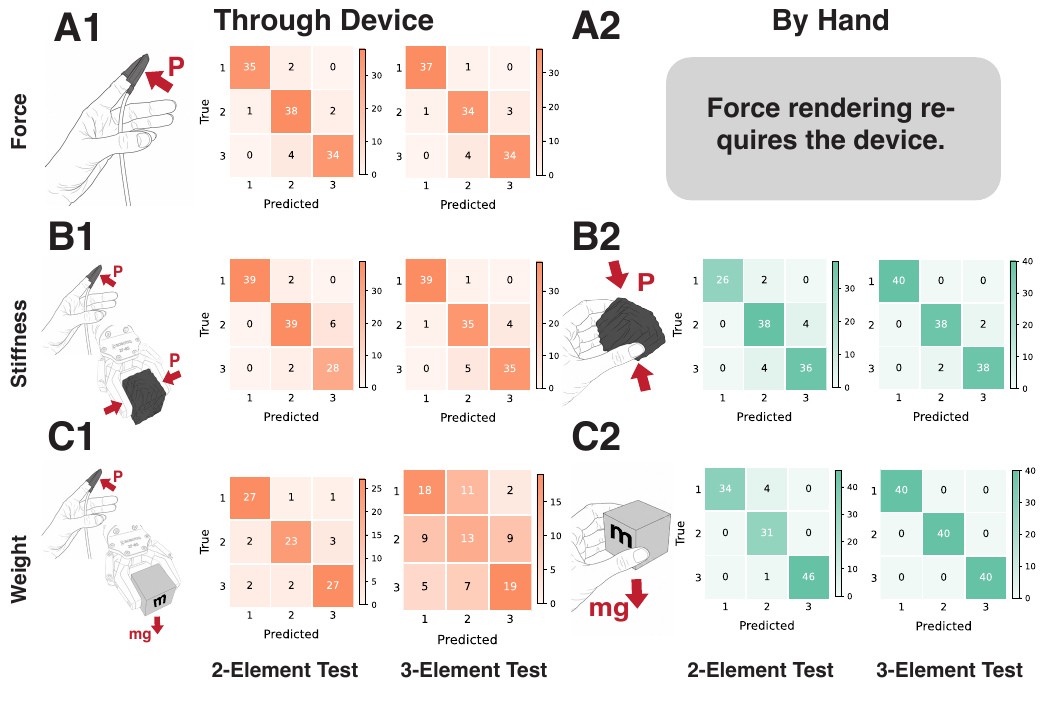}
    \caption{\textbf{Perception Study Results} Confusion matrices for increasing force (\textbf{A1}; 1, 2, 3 correspond to 150, 300, and 450 mbar), stiffness (\textbf{B1} and \textbf{B2}; 1, 2, 3 correspond to 0.630 N/mm, 1.77 N/mm, and 3.35 N/mm), and weight discrimination tests (\textbf{C1} and \textbf{C2}; 1, 2, 3 correspond to 150~g, 300~g, and 450~g), ordering objects in sets of 2 and 3; counts on the diagonal show correct guesses; these counts are used to estimate the accuracy for each task; (\textbf{A2}) force results are shown only for the Haptics case since the test was only performed using the device.}
    \label{fig:confusionmatrices}
\end{figure}

To evaluate the interpretability of the haptic feedback provided by the device, we conducted a series of perceptual discrimination tasks probing force, stiffness, and weight with $n=25$ participants. Participants were recruited through university email lists and Slack channels. Participants provided informed consent, and the protocol received approval from the Stanford University Institutional Review board (\#65022). These tasks were designed to progressively increase in complexity, moving from direct haptic feedback alone to mediated interaction through teleoperation. 
Teleoperation was performed using a Kinova Gen3 7 DoF arm and a Robotiq 2F-85 gripper with a textile actuator on the inside of each finger for sensing, and a corresponding textile actuator on the index and thumb pads of the participants for haptic feedback. The robot position was controlled using wrist and finger (thumb and index) tracking on the right or left arm through a Microsoft HoloLens 2 worn by the participant (see Methods section and Fig.~\ref{fig:teleop_overview} for more detail). 
The results for the five tasks are illustrated with confusion matrices in Fig.~\ref{fig:confusionmatrices}. 

\subsubsection*{Force}
We first assessed force discrimination using the haptic device alone, without involvement of the robotic manipulator. Participants were asked to discriminate between discrete force levels delivered directly to the index and thumb fingertips via the haptic device. The three levels of force were 150, 300, and 450 mbar.

Participants demonstrated consistently high accuracy in this task. In the 2-element force discrimination condition (Fig.~\ref{fig:confusionmatrices}A1), participants correctly identified whether the second force was stronger, weaker, or equal to the first with an accuracy of 92.2\%. Performance remained stable when the task was extended to 3-element comparisons, with an accuracy of 92.1\% (Fig.~\ref{fig:confusionmatrices}A1). These results indicate that the pressure-based fingertip feedback enabled reliable perception of force magnitude even as comparison complexity increased.

\subsubsection*{Stiffness}
We next evaluated stiffness discrimination under two conditions: direct manual interaction with three soft cubes, and haptic feedback mediated by the robotic gripper. The three cube stiffnesses were 0.630, 1.77, and 3.35~N/mm. In the mediated condition, the gripper closed onto each cube with a fixed closing motion while participants faced away from the setup to prevent visual cues, and the gripper sensors drove the fingertip pouches directly.

Participants performed comparably in the two conditions (Fig.~\ref{fig:confusionmatrices}B1 and B2). In the 2-element task, accuracy was 90.1\% by hand and 91.4\% through the device. In the 3-element task, it was 96.7\% by hand and 90.8\% through the device, a difference of 5.9 percentage points. That mediated feedback matched direct touch on the 2-element task and came within six points on the harder 3-element task, indicating that fingertip indentation feedback preserved stiffness information with high fidelity, supporting the use of pressure-based cues for conveying stiffness.

\subsubsection*{Weight}
Finally, we examined weight discrimination, again comparing direct manual interaction to haptic feedback mediated through teleoperation. The three weights were rigid blocks of 150, 300, and 450~g. In the teleoperated condition, participants actively controlled robot position and gripper opening by moving their right wrist and their index finger and thumb, respectively, and were asked to pick up each weight with the gripper and judge it from the haptic feedback at their fingertips.

Weight discrimination diverged sharply between the two conditions (Fig.~\ref{fig:confusionmatrices}C1 and C2). In the 2-element task, accuracy was 95.7\% by hand and 87.5\% through teleoperation, a difference of 8.2 percentage points. In the 3-element task, it was 100\% by hand and 53.8\% through teleoperation. Mediated feedback therefore tracked direct lifting closely when only two weights had to be ordered, but lost most of that performance once a third was added. The added cognitive load and motor complexity of the weight discrimination task appeared to challenge participants' ability to reliably encode and recall the associated haptic sensations across multiple comparisons, unlike the force and stiffness discrimination tasks.

The order in which force, stiffness, and weight were displayed in the 2- and 3-element conditions was randomly generated in a test protocol sheet for each participant.

\subsection*{Improved teleoperation performance with reduced mental load}

\begin{figure}
    \centering
    \includegraphics[width=\linewidth]{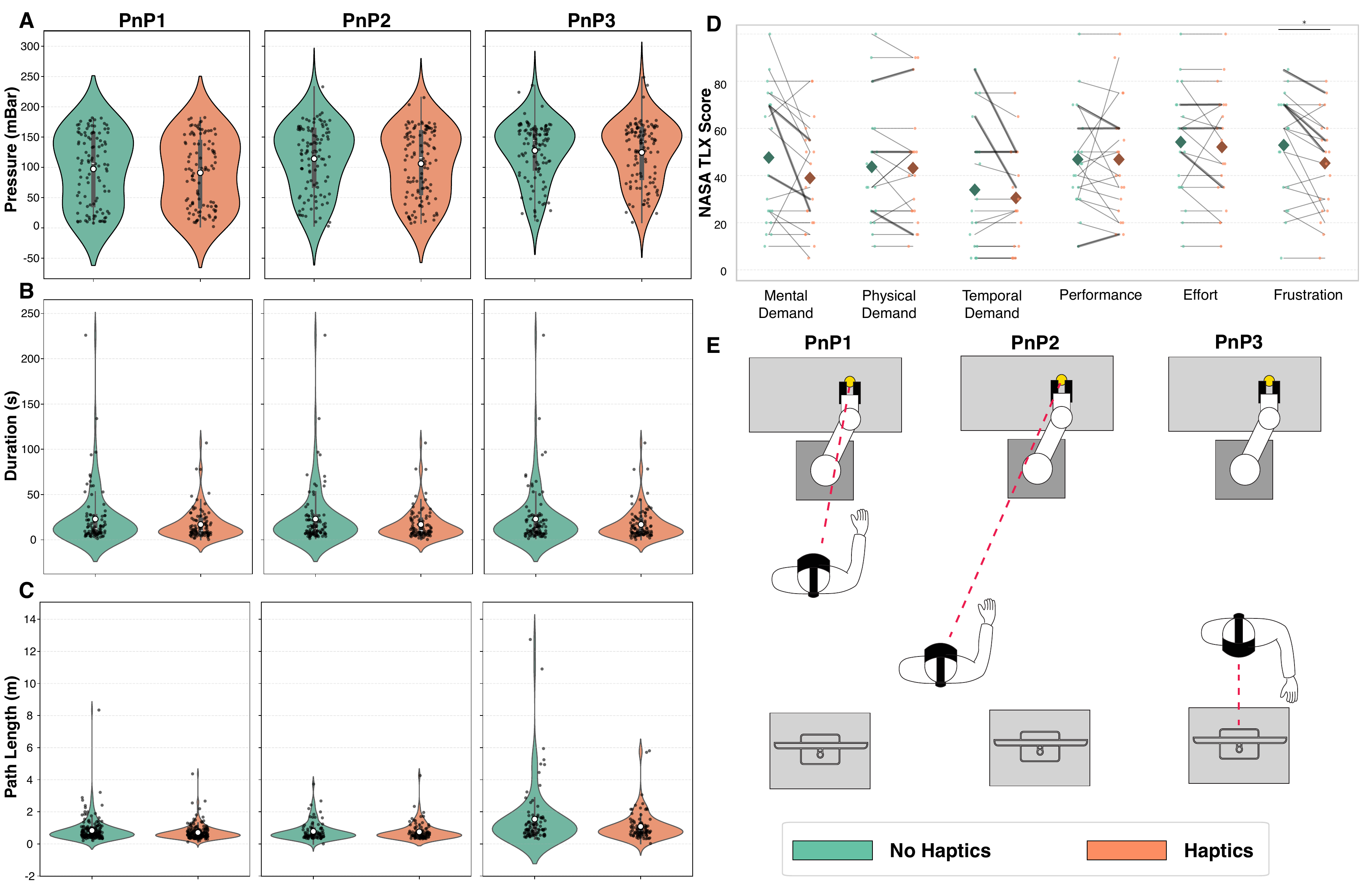}
    \caption{\textbf{Quantitative and Qualitative Teleoperation Results} \textbf{A} Pressure, \textbf{B} Duration, \textbf{C} Path length results for the three pick-and-place tasks (PnP1: direct view, PnP2: obstructed view, PnP3: camera view) for haptics and no haptics; the violin plots show the mean (white dot), the inter-quartile range (grey bar), and the full range (grey line); \textbf{D} NASA Task Load Index Scores: Participants reported perceived workload across six dimensions, rated from 0 (none) to 100 (extreme); bolded lines indicate the three slowest participants (longest task duration); asterisks denote statistically significant differences between conditions (p $<$ 0.05, n = 19); \textbf{E} Test setup for the three pick-and-place tasks, where participants faced the table and robot in PnP1 and PnP2, and faced a computer monitor that displayed the wrist camera feed from the robot in PnP3. }
    \label{fig:ViolinPlots_Results}
\end{figure}

We next evaluated whether the haptic feedback provided by the device translated into measurable improvements during teleoperated manipulation tasks. Participants performed three pick-and-place (PnP) tasks under visual-only and haptic feedback conditions: PnP1 in which they had direct view of the task, PnP2 in which they had an obstructed view of the task, and PnP3 in which they viewed only the camera feed from the robot (see Fig.~\ref{fig:ViolinPlots_Results}E). For all tasks, participants were asked to pick up a hackey~sack from a starting position and place it at another position marked by tape. Participants were encouraged during practice trials prior to PnP1 to use the minimum grip force required to reliably grip the object. Within each task, the order of haptics vs no-haptics conditions were pseudo-randomized for each participant. The quantitative results are shown in Fig.~\ref{fig:ViolinPlots_Results}A, B, and C respectively in the form of violin plots. Teleoperation results are shown for $n=19$ participants due to exclusions stemming from device malfunctions during the study.

Across all three tasks, participants consistently applied less pressure when fingertip haptic feedback was available. In PnP1, average applied pressure decreased by 19.2\% relative to the visual only condition. In PnP2, pressure decreased by 23.1\%, and in PnP3 by 7.2\%. These reductions indicate that haptic feedback enabled gentler interaction with the delicate object across task conditions, even when visual information was present.

Task efficiency also improved under haptic feedback in two of the three conditions. Participants completed PnP1 and PnP3 21.8\% and 25.6\% faster, respectively, while following trajectories that were 14.6\% and 28.9\% shorter. Speed improvements in PnP1 and PnP3 were attributed to faster object transport, whereas speed of initiating a grasp remained unchanged (see Fig.~S8 for PnP time allocation by condition). In contrast, overall improvements in PnP2 were minimal, with path length and average completion time remaining largely unchanged. In PnP2, transport performance gains were washed out by slower grasp initiation. Despite being rated by participants as the most challenging task, performance gains in PnP2 may have been limited by increased familiarity with the teleoperation interface by this stage of the study.

Subjective workload assessments further reflected the impact of haptic feedback. After the pick-and-place tasks in each condition, participants also completed the NASA Task Load Index (TLX), which rates perceived workload across six subscales from 0 (none) to 100 (extreme). As shown in Fig.~\ref{fig:ViolinPlots_Results}D, TLX scores decreased across four of the six subscales when haptic feedback was provided, including Mental Demand, Temporal Demand, Effort, and Frustration. Mean Mental Demand decreased from 47.6 to 39.0 (-8.7 points, 18.1\%), and Frustration decreased from 52.9 to 45.3 (-7.6 points, 14.4\%), the latter being statistically significant. Reductions in Temporal Demand, and Effort were more modest (2 to 3 points). Perceived performance and Physical Demand showed little to no difference between conditions (46.8 vs 46.8 and 43.7 vs. 43.2, respectively). 

Because perceived workload is often correlated with task duration, we additionally highlight the three participants with the longest task completion times; shown in bold in Fig.~\ref{fig:ViolinPlots_Results}D (additional individual participant level results can be found in Figs.~S5 to S7). For these participants, reductions in subjective workload under the haptic condition were more pronounced than the group average: Mean Mental Demand decreased from 63.3 to 41.7 (-21.7 points, 34.3\%), Performance improved from 58.3 to 36.7 (-21.7 points, 37.2\%), and Frustration decreased from 61.7 to 43.3 (-18.3 points, 29.7\%). These results suggest that haptic feedback is particularly beneficial for longer or more complex tasks.

After completing the pick-and-place tasks in each condition, participants rated the system using the standardized \gls{sus}, a ten-item questionnaire that yields a single usability score from 0 to 100. Their responses suggested improved perceived usability when haptic feedback was provided. Mean \gls{sus} scores increased from 61.1 $\pm$ 14.6 in the No Haptics condition to 64.9 $\pm$ 16.3 in the Haptics condition, corresponding to a 3.8-point improvement. Although this difference did not reach statistical significance ($p = 0.090, d_z=-0.41$), the observed trend indicates that haptic feedback may enhance users’ overall perception of system usability. In open-ended comments, several participants also reported perceiving delay during operation, attributing latency of the robotic arm to the haptic feedback. 

A detailed report of the statistical analysis is provided in the Supplemental Information.

\section*{Discussion}
The central contribution of this work is a design in which tactile sensing and haptic display use the same physical mechanism, regulated in closed loop and carried entirely on the body. Prior wearable pneumatic devices have had to choose between these properties: those achieving stable, sustained pressure rely on external compressors and valve banks that confine them to fixed setups, while untethered designs rely on open-loop solenoid actuation, which limits feedback to transient pulses rather than the regulated, continuously varying contact forces teleoperation requires. Distributing a proportional valve, pressure sensor, and microcontroller to each channel resolves that tradeoff: regulation happens locally, so the device scales across channels without a central bottleneck and stays light enough to wear. And because the sensing and actuation use the same modality, no information is lost between the robot and the operator.

The device characterization validates this design. Closed-loop pressure regulation, high flow capacity, and a rapid, well-damped step response enable stable, continuous fingertip indentation at perceptually relevant force and temporal scales. Unlike many prior pneumatic haptic systems that rely on open-loop or transient actuation, our device maintains regulated pressure over extended periods without rapid decay. The distributed control architecture allows individual channels to be regulated locally while remaining synchronized across the device, supporting low-latency response and sustained interaction. These characteristics are critical for tasks involving prolonged contact and fine force regulation, particularly in teleoperation settings where force cues must remain stable and consistent over time. 

Across force and stiffness discrimination tasks, participants demonstrated high accuracy when using the haptic feedback device. Stiffness discrimination performance in particular was close to that of direct manual interaction, indicating that the device preserves compliance information with high fidelity. These findings support the use of indentation-based cues for conveying continuous contact information, such as force magnitude and material properties, and suggest that such cues can be interpreted reliably without extensive training. In contrast to vibration-based feedback, pressure-based indentation provides a direct mechanical mapping to contact forces, which likely contributes to its perceptual robustness and interpretability.

Weight discrimination revealed a more nuanced picture. While participants were able to discriminate object weight with near-perfect accuracy during direct manual interaction, performance declined in the harder teleoperated condition, particularly for 3-element comparisons. This reduction highlights the role of task-level cognitive demands in shaping perceptual performance. In the teleoperated setting, participants were required to integrate fingertip pressure cues with simultaneous control of arm motion and grip force, increasing memory and coordination requirements. These results emphasize that perceptual accuracy depends not only on the fidelity of the pressure signal, but also on how directly that signal corresponds to the property being judged while the participant is simultaneously controlling the robot.

The twin-system configuration further extends the capabilities of the device by enabling bidirectional tactile communication. By pairing two identical devices, sensed pressure signals can be transmitted and reproduced in real time without intermediate abstraction. Beyond enabling haptic mirroring, this architecture allows tactile interaction data to be recorded using the same sensing modality under both haptic and non-haptic conditions. In the present study, this made it possible to capture synchronized tactile pressure data during teleoperated manipulation regardless of feedback condition. Such paired datasets provide a foundation for treating tactile pressure as a first-class modality in learning from demonstration, where policies can be trained to incorporate contact information alongside vision and proprioception.

When deployed in teleoperated manipulation tasks as a validation context, the device led to consistent reductions in applied force and improvements in task efficiency, particularly under visually constrained conditions. Participants also reported lower mental demand and frustration when haptic feedback was available, indicating that tactile cues reduced the cognitive effort required to infer contact state and regulate interaction forces. Consistent with the framing in the introduction, teleoperation is not presented here as the primary application of the device. Rather, it serves as a demonstrative use case illustrating how the device facilitates perceptual interpretability and bidirectional feedback within a human-robot interaction context.

Prior work has shown that haptic feedback can reduce average task completion time and substantially increase the number of usable demonstrations collected during teleoperation. In particular, vibrotactile feedback has been reported to improve efficiency in aggregate, although these time advantages largely diminish when analysis is restricted to curated successful demonstrations~\cite{Cuan2024}. In contrast, our results show that pressure-based fingertip feedback yields persistent improvements in task efficiency and force regulation even under visually constrained conditions, alongside significant reductions in subjective mental demand. This suggests that closed-loop, indentation-based feedback may not only increase data yield, but also improve the quality and interpretability of individual demonstrations.

This study has several limitations. The experimental setup relied on multiple subsystems for robot control, visual feedback, and haptic feedback, which occasionally resulted in technical interruptions and participant exclusions. Some participants also conflated latency in the robotic system with latency in the haptic device, underscoring the need for tighter system integration in future studies. From a hardware perspective, further reductions in size, improvements in force resolution, and continued refinement of control strategies remain important directions for development.

Overall, this work demonstrates that closed-loop, pressure-based fingertip haptic feedback can support reliable tactile discrimination while reducing cognitive load during interaction. By unifying tactile sensing and haptic feedback within a single pneumatic platform and enabling bidirectional tactile communication, the device provides a practical foundation for future research in tactile interaction, teleoperation, and data-driven learning from touch.




\section*{METHODS}

\subsection*{Pneumatic Controller}

\begin{figure} [h]
    \centering
    \includegraphics[width=0.9\linewidth]{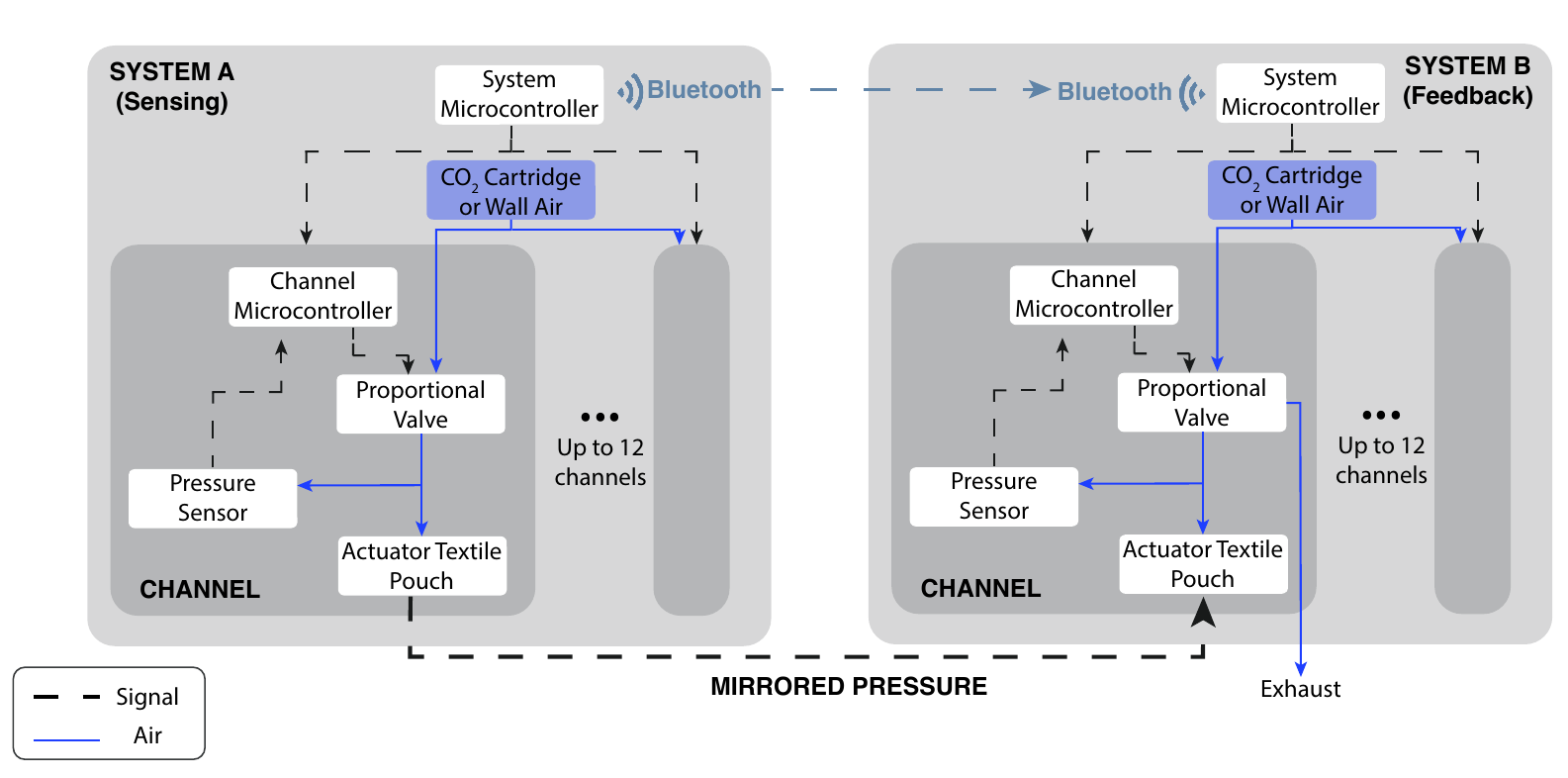}
    \caption{\textbf{System architecture of the closed-loop pneumatic haptic controller} Each  channel uses a proportional valve, pressure sensor, and local microcontroller to enable independent closed-loop pressure regulation of textile pneumatic actuators. Multiple channels are coordinated by a system-level microcontroller, which also handles wireless communication. A single CO2 cartridge or wall air supports all channels in a system. Two identical systems can be paired to transmit sensed pressure signals (System A) and reproduce them in real time (System B), enabling bidirectional tactile communication.}
    \label{fig:systemarchitecture}
\end{figure}

\subsubsection*{Hardware Architecture}

The pneumatic controller was designed as a compact, untethered system capable of closed-loop pressure regulation for multiple soft actuators. As shown in Fig.~\ref{fig:systemarchitecture}, each actuator channel consists of a proportional piezoelectric valve, an pressure sensor, and a dedicated microcontroller, forming a local actuator–sensor–controller unit. This distributed architecture allows each channel to regulate pressure independently while remaining synchronized with the rest of the device. The base weight of the device is 204 g, with an additional 52g per channel (so for a two-channel device, 308 g total). For the study, we use two paired devices: one connected to the robot, and one to the human operator.

Compressed air is supplied either by an external source or an onboard reservoir and routed through the piezoelectric valves to the textile actuators. Pressure sensors are placed proximal to each actuator to enable accurate local feedback. The controller electronics are housed in a lightweight enclosure and include power regulation, wireless communication hardware, and safety circuitry. The device was designed to minimize pneumatic dead volume and leakage in order to support sustained pressure output during prolonged interaction.

\subsubsection*{Software Architecture}

Closed-loop pressure control is implemented locally on each microcontroller using proportional\-integral control. Pressure sensor readings are sampled continuously and used to regulate valve opening in real time, allowing each channel to track desired pressure setpoints with low latency and minimal overshoot. Control parameters are identical across channels, enabling scalable operation with multiple actuators.

A central communication layer coordinates setpoint updates and data logging across channels. Wireless communication is used to transmit pressure commands and sensor data between paired devices in the twin-system configuration. This architecture supports both local pressure regulation and bidirectional tactile communication while avoiding centralized control bottlenecks. All firmware was developed to prioritize deterministic timing and robustness under continuous operation.

\subsection*{Teleoperation System Overview}
\begin{figure}
    \centering
    \includegraphics[width=1.0\linewidth]{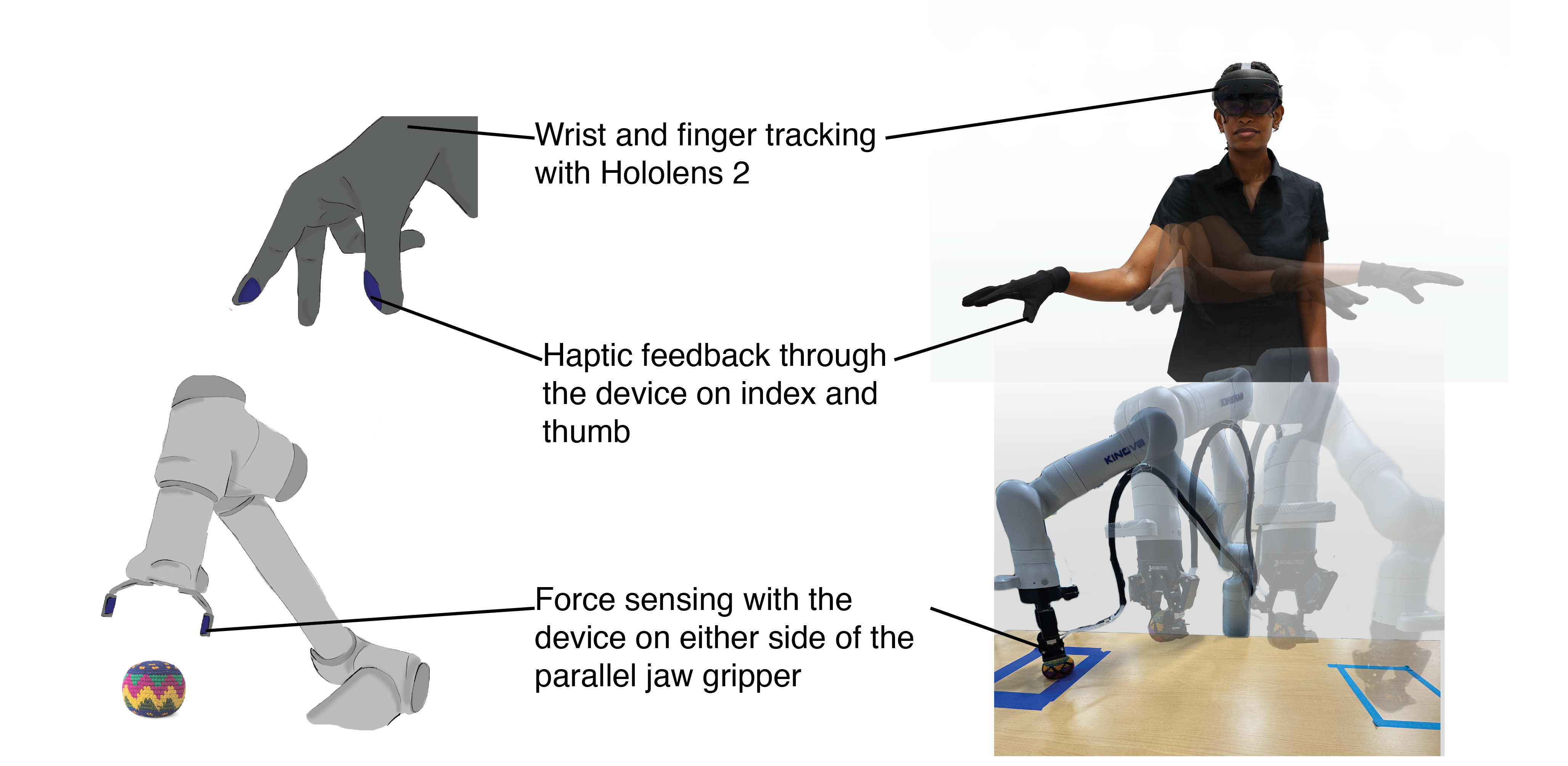}
    \caption{Overview of the hand tracking-based teleoperation system; finger and wrist tracking through the HoloLens 2; haptic feedback to the index and thumb with one twinned device, force sensing on either side of the parallel-jaw gripper with the other twinned device.}
    \label{fig:teleop_overview}
\end{figure}

\subsubsection*{Haptic Feedback System}

Haptic feedback was delivered to the user through textile-based pneumatic actuators mounted on the thumb and index fingertips, replicating pressure variations corresponding to contact forces on either side of the parallel-jaw gripper (see Fig.~\ref{fig:teleop_overview}). The actuators were independently regulated by the closed-loop pneumatic controller described above, allowing pressure to be modulated continuously during interaction. Actuators were positioned such that applied pressure produced normal indentation at the fingertip pad, directly conveying contact force information. The actuators were fabricated from two layers of 70 Denier TPU coated Ripstop nylon, heat-pressed using a stencil, and a barb was then inserted and secured with glue. The detailed manufacturing method is reported by Coram et al.~\cite{Coram2024SealingActuators}.

During teleoperation, pressure commands for the haptic actuators were generated from force measurements acquired at the gripper. Normal forces measured at the gripper were mapped to actuator pressure setpoints using a linear scaling factor of four, determined empirically to ensure perceptible yet comfortable indentation. Pressure setpoints were updated continuously and tracked locally by the controller, enabling sustained force feedback without perceptible drift or decay.

\subsubsection*{Robot Control}

Participants controlled the Kinova Gen3 7 DoF arm using direct hand tracking through a Microsoft HoloLens 2, without holding or manipulating any physical input devices. This design kept the hands and fingers fully unencumbered, allowing haptic feedback to be delivered directly to the fingertips. Wrist motion was mapped to Cartesian velocity commands, while thumb-index finger distance controlled the opening of the parallel-jaw gripper. This approach enabled continuous, intuitive control of both arm motion and grasp without introducing additional physical interfaces that could interfere with tactile perception. HoloLens 2 hand tracking has been independently characterized against a Vicon motion capture reference, with mean fingertip position errors of 2 to 4 mm and thumb-index pinch span errors below 1 mm~\cite{Bertolasi2025}, sufficient for the velocity-based control mapping used here. The Gen3 has a manufacturer-specified position repeatability of 1.0 mm (2$\sigma$) in its 7 DoF configuration, valid for calibrated robots in a fixed-base installation such as the one used here~\cite{KinovaGen3Specs}. Full implementation details are provided in the Supplemental Information.

\subsection*{Hypotheses}

Based on the design and capabilities of the device, we tested the following hypotheses:

\begin{enumerate}
    \item The pneumatic fingertip haptic feedback system enables discrimination of force and object properties, including stiffness and weight.
    \item During teleoperation, the inclusion of haptic feedback reduces the forces transmitted to the manipulated object.
    \item During teleoperation, haptic feedback improves task performance, reflected in reduced completion time and increased efficiency.
    \item Haptic feedback lowers cognitive load during teleoperation.
    \item The use of textile-based sensors supports comfort and overall usability.
\end{enumerate}

\subsection*{User Study Procedure}

Participants were recruited to evaluate the system under conditions with and without haptic feedback. The study protocol was approved by the Stanford Institutional Review Board (Protocol 65022). Twenty-five able-bodied participants completed the study (10 female, 14 male, 1 non-binary), ranging in age from 19 to 67 years (M = 31, SD = 11.7). Twenty-two participants were right-handed and three were left-handed. All participants provided informed consent prior to participation. The study lasted approximately 1.5 hours per participant and consisted of two phases: Perception and Teleoperation.

\subsubsection*{Perception Tasks}

The perception phase consisted of five tasks designed to assess participants’ ability to interpret fingertip haptic feedback across increasing levels of complexity. Tasks progressed from direct haptic stimulation alone to robot-mediated interaction and teleoperation. The five tasks were:

\begin{enumerate}
    \item \textbf{Direct force discrimination}: Participants discriminated between discrete levels of fingertip pressure delivered directly by the haptic device, without involvement of the robotic manipulator.
    \item \textbf{Direct stiffness discrimination}: Participants manually squeezed physical objects of differing stiffness to establish a baseline for human stiffness perception.
    \item \textbf{Haptic-mediated stiffness discrimination}: The robotic gripper interacted with objects of differing stiffness using a fixed grasp motion, and the resulting contact forces were conveyed to participants via fingertip haptic feedback.
    \item \textbf{Direct weight discrimination}: Participants lifted visually identical objects of different mass\-es by hand to establish a baseline for weight perception.
    \item \textbf{Teleoperated weight discrimination}: Participants controlled the robotic arm and gripper to lift objects of different masses, relying on fingertip haptic feedback to discriminate weight during teleoperation.
\end{enumerate}

Across all five tasks, participants were seated to eliminate whole-body motion cues, and visual information was restricted as appropriate to isolate haptic perception. Full task parameters and trial structures are provided in the Supplemental Information.

\subsubsection*{Teleoperation Training}

Before formal evaluation, participants completed a five-minute training session to familiarize themselves with the teleoperation interface and haptic feedback. The training duration was selected in accordance with the researchers’ domain expertise and established practices from prior teleoperation studies. During training, participants practiced manipulating an empty aluminum soda can, a deliberately fragile object that encouraged careful force regulation and precise motion. Participants were instructed to move the can around the workspace without deforming it. Half of the training period was performed with haptic feedback enabled and half without haptic feedback, with the order randomized across participants.

\subsubsection*{Teleoperation Evaluation Tasks}

The teleoperation phase evaluated whether fingertip haptic feedback improved manipulation performance, force regulation, and perceived workload under varying visual conditions. Participants completed a series of pick-and-place tasks of increasing visual difficulty while controlling the robotic arm and gripper. In all tasks, participants were instructed to pick up a soft object from a fixed starting location and place it at a fixed goal location marked with tape on the table.

A hackey~sack (round, sand-filled cloth bag) was selected as the target object because its compliant structure makes improper grasping difficult to assess visually, while remaining sufficiently stiff to generate reliable force signals within the sensitivity range of the pneumatic tactile sensors. This choice ensured that successful manipulation depended primarily on force regulation rather than visual deformation cues.

The three teleoperation tasks were:

\begin{enumerate}
    \item \textbf{Pick-and-place with direct view (PnP1)}: Participants performed the task while standing behind the robot with an unobstructed view of the workspace, gripper, and object.
    \item \textbf{Pick-and-place with obstructed view (PnP2)}: Participants performed the same task while standing farther from the robot, where portions of the workspace were intermittently occluded by the robot itself.
    \item \textbf{Pick-and-place with robot-mounted camera view (PnP3)}: Participants relied solely on a wrist-mounted camera feed displayed on a monitor, with no direct line of sight to the robot or object.
\end{enumerate}

As shown in Fig~\ref{fig:ViolinPlots_Results}E, each task was performed under both visual-only and haptic-feedback conditions in a counterbalanced order. PnP1 included a larger number of repetitions to capture steady-state performance, while PnP2 and PnP3 emphasized increased visual uncertainty. The tasks were intentionally simple to isolate the effects of haptic feedback on applied force, task efficiency, and perceived workload rather than on task planning complexity. Detailed trial counts, training procedures, and performance metrics are provided in the Supplemental Information.

\subsubsection*{Metrics}

Both quantitative and qualitative metrics were collected. Quantitative measures included fingertip pressure signals, task completion time, and gripper path length. Qualitative measures included the System Usability Scale (SUS) and the NASA Task Load Index (TLX), which assessed perceived usability and workload, respectively. Detailed definitions of all metrics and data processing procedures are provided in the Supplemental Information.

\subsubsection*{Statistical Analyses}

Statistical analyses were performed in Python using the SciPy statistics package. Paired t-tests were used to compare conditions with and without haptic feedback for all measured variables. Trials with missing data due to logging errors were excluded symmetrically across conditions.

\newpage


\section*{RESOURCE AVAILABILITY}


\subsection*{Lead contact}


Requests for further information and resources should be directed to and will be fulfilled by the lead contact, Cosima du Pasquier (cosimad@stanford.edu).

\subsection*{Materials availability}



\subsection*{Data and code availability}

\begin{itemize}
    \item All data reported in this paper will be shared by the lead contact upon request.
    \item The device control software is available as the open-source Python package \texttt{piezense} on PyPI: \url{https://pypi.org/project/piezense/}.
    \item Any additional information required to reanalyze the data reported in this paper is available from the lead contact upon request.
\end{itemize}



\section*{ACKNOWLEDGMENTS}



This work was supported in part by the National Science Foundation under Grant No. 2301355. Aliyah Smith was supported by NSF Graduate Research Fellowship
DGE-1656518. The authors thank the Stanford Robotics Center for their support. 

\section*{AUTHOR CONTRIBUTIONS}


Conceptualization, C.d.P., A.S., and A.M.O.; methodology, C.d.P., A.S., and A.M.O.; investigation, C.d.P., A.S., S.H., J.P., I.A.C., and A.C.; writing-–original draft, C.d.P. and A.S.; writing-–review \& editing, C.d.P, A.S., A.C., M.K., A.M.O.; funding acquisition, C.d.P., M.K., and A.M.O.; resources, M.K., and A.M.O.; supervision, C.d.P., M.K., and A.M.O.

\section*{DECLARATION OF INTERESTS}



C.d.P. is a co-founder, officer, and shareholder of Haptica, Inc., which is commercializing technology related to this work. A.M.O. serves as a scientific advisor to and holds an equity interest in Haptica, Inc. J.P. subsequently provided paid contract engineering services to Haptica, Inc., beginning after the work reported here was completed. C.d.P. and S.H. have a provisional patent on technology related to this work (submission \#S25-344-PROV). The remaining authors declare no competing interests.

\section*{DECLARATION OF GENERATIVE AI AND AI-ASSISTED TECHNOLOGIES}


During the preparation of this work, the author(s) used ChatGPT (OpenAI) and Claude (Anthropic) in order to enhance the clarity and readability of the written text. After using this tool or service, the author(s) reviewed and edited the content as needed and take(s) full responsibility for the content of the publication.

\section*{SUPPLEMENTAL INFORMATION INDEX}




\begin{description}
\item Document S1. Supplemental methods, Figures S1-S9 and Tables S1-S7.
\end{description}

\bibliography{references}

\clearpage


\section*{Supplementary material}

**Supplemental Information**

# A Wearable Pneumatic Device for Continuous, Closed-Loop, Bidirectional Tactile Interaction

C. du Pasquier, A. Smith, S. Huber, J. Phelps, I. Cohen, A. Chen,
M. Kennedy III, A. M. Okamura

## 1 Controller Design and Architecture

### 1.1 System architecture and key components

Table S1: **Table S1. Key system components and interfaces.**

| Subsystem | Component and identifier |
|---|---|
| Channel microcontroller | Nordic Semiconductor nRF5340 |
| Pressure sensor | TE Connectivity MS583702BA |
| Proportional valve | Festo VEMP-BS-3-13-D19-F-28T1 piezo valve |
| Wireless | BLE |
| Analog outputs | Two DAC channels, 0 to 10 V |
| High voltage drive | 0 to 310 V piezo drive stage |
| Valve pressure rating | 0 to 2.7 bar |

Each device is organized as a set of up to twelve identical actuator channels coordinated by a system microcontroller. Each channel comprises a proportional piezo valve, a pressure sensor, and a channel microcontroller that runs local closed loop pressure control (performing proportional-integral control in software). Two identical devices can be paired so that one device senses pressure signals and transmits them wirelessly to the other device, which reproduces the received pressure signals in real time.

**Electronics and sensing** Each channel uses a Nordic Semiconductor nRF5340 microcontroller and a TE Connectivity pressure sensor (MS583702BA). The pneumatic control valves used in this study are Festo VEMP piezo proportional valves, which support continuous actuation over the full pressure range (0–1.7 bar) and nominal flow rates up to 28 L/min. These valves use integrated piezo actuators for proportional control, enabling graded pressure modulation rather than simple digital switching [1]. The schematic architecture includes a Leader PCB and a Control PCB. The device uses BLE for wireless communication. The architecture includes DC to DC converters and two DAC outputs per channel. The DAC outputs provide 0 to 10 V signals that drive a high voltage stage (0 to 310 V) used for piezo actuation. The valve and pneumatic pathway are rated for 0 to 2.7 bar.

**Power and rails** The schematic indicates 24 V and 3.3 V system rails, plus intermediate rails (12 V and 16 V) used for analog and high voltage generation. The DAC and sensor communicate over I2C. The device is powered by a OVONIC 130C 6S 1100mah 22.2 V Lipo Battery.

**Pneumatics** Compressed air is supplied either by wall air or by an onboard CO2 cartridge regulator. In the airtightness protocol, the inlet port of the VEMP valve is fully opened to allow

unrestricted airflow from the CO2 cartridge regulator to the actuators, then the regulator is closed to isolate the device and prevent additional air intake. We use 2 mm ID, 4 mm OD silicone tubing and plastic barbed fitting connectors.

## 1.2 Local closed loop control and software architecture

Each channel microcontroller runs closed loop pressure control using the collocated pressure sensor. This distributed control approach avoids centralized control bottlenecks and supports scaling to multiple channels.

**Controller implementation** The closed-loop pressure controller runs on the nRF5340 microcontroller of each Control PCB under the Zephyr RTOS. The pressure sensor thread samples at approximately 500 Hz (1 ms sleep interval plus two 500 µs ADC conversions per reading). The control loop runs free-running with no fixed period; elapsed time $dt$ is computed dynamically from the Zephyr system clock (`k_uptime_get()`), and the effective control update rate is bounded in practice by the sensor update rate.

Feedback is computed from a rolling mean of the 8 most recent pressure readings, which attenuates high-frequency sensor noise. The error signal is defined as $\epsilon(t) = P_0(t) - P_{\text{rolling window}}(t)$, where $P_0(t)$ is the pressure setpoint provided by the Leader PCB. A symmetric deadband of $\pm 2$ mbar suppresses actuation for small errors; when $|\epsilon(t)| < 2$ mbar the controller output is forced to zero and no DAC write is issued.

Due to the asymmetric dynamic response of the inlet and exhaust piezo valve channels, separate controllers are tuned for each path. When $\epsilon(t) \geq 2$ mbar, a PI controller ($K_p = 0.6$, $K_i = 0.25$, $K_d = 0$) drives the inlet DAC; when $\epsilon(t) \leq -2$ mbar, a PID controller ($K_p = 0.4$, $K_i = 0.5$, $K_d = 0.5$) drives the exhaust DAC. Both controllers were tuned manually by iterative adjustment on the physical device. DAC output voltages are clamped to the 0–10 V rated input range of the Festo VEMP piezo valve driver.

Anti-windup is implemented through four combined mechanisms: (1) the integral term is zeroed whenever the error crosses zero (sign-change reset); (2) the integral is hard-clamped to $\pm 400$; (3) a per-cycle decay factor of 0.9999 is applied as a soft leak to prevent slow accumulation; and (4) when the setpoint changes by more than 50 mbar the integral is hard-reset to zero, while smaller setpoint changes scale the integral by 0.95.

**Host GUI role in characterization and twin-system testing** In the twin-system delay test, pressure data are transmitted via Bluetooth from the Leader PCB of the sensing device to a host computer, through which the signal is logged and forwarded using a python script to the second device. The second device actuates the corresponding pressure response and reports status back to the computer, enabling timing measurement.

## 1.3 Actuator characterization protocols

### 1.3.1 Blocked force and force bandwidth

The objective of this test was to quantify two critical performance metrics of the textile actuator: its maximum output force and its frequency-dependent force response, known as the force bandwidth. These parameters directly influence the actuator's effectiveness in haptic feedback applications. The maximum output force determines the strength of the tactile sensations that the actuator can deliver, which is essential for perceptibility through the skin. The force bandwidth reflects how quickly and accurately the actuator can respond to dynamic input signals, an important requirement for

conveying realistic, time-varying haptic cues.

**Methods** The test setup shown in Fig. S1a and S1b, consists of a custom-built force measurement platform. A 3D printed frame similar to the one use by du Pasquier et al. [2] was mounted on a rigid steel base to ensure stability. A circular cutout in the frame houses the textile actuator. Below the actuator, a calibrated force sensor (ATI Nano-17) [3] measured the downward force generated during actuation.
To constrain the actuator and direct its expansion force toward the sensor, a rigid Plexiglass plate was fixed on top of the actuator. This blocked force configuration prevented vertical displacement and ensured that the pressurization force was directed into the force sensor.
To assess the actuators' peak force at 2600 mbar input pressure, static force tests were conducted on ($n = 3$) textile actuators.
In the second phase of the test, the actuators were repeatedly fully pressurized (opening the proportional valve completely) and fully depressurized (closing the valve completely) at using a frequency sweep from 1 to 100 Hz.

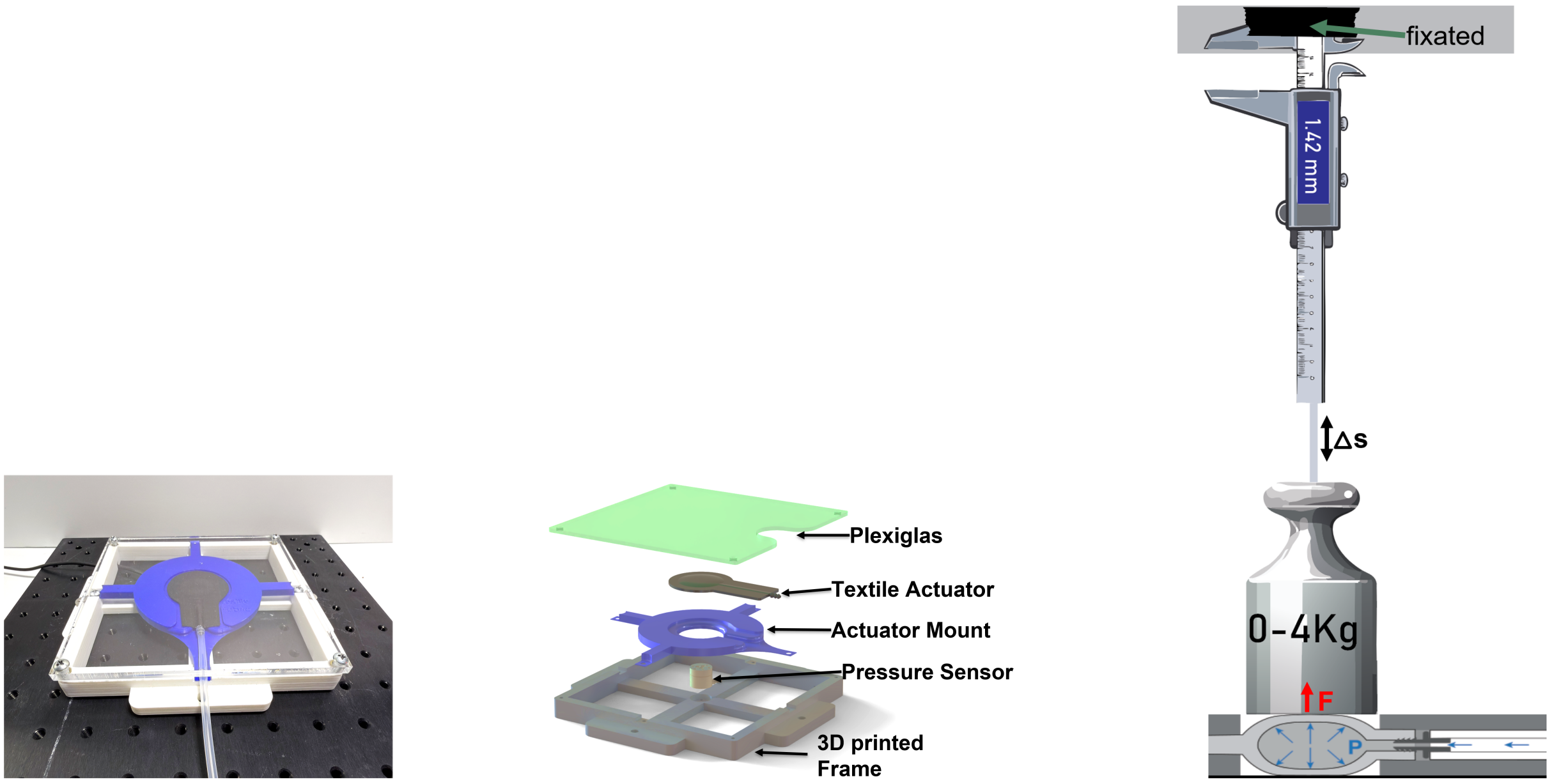


(a) (b) (c)

Figure S1: Experimental setups for actuator characterization: (a) assembled force test rig, (b) schematic of force test, and (c) displacement test rig.

**Results** At a peak input pressure of 2650 mbar, the actuators produce an average maximum force of 76.2 N (±2.8 N SD).
The force bandwidth results are shown in Fig. S2. The magnitude is normalized to 40 N (0 dB reference). The actuators maintain on average a steady response up to 34.2 Hz (±1.2 Hz SD). In contrast, actuators controlled using Airport reaches its point 3 dB only at 14.5 Hz [2], indicating a substantially lower force bandwidth.

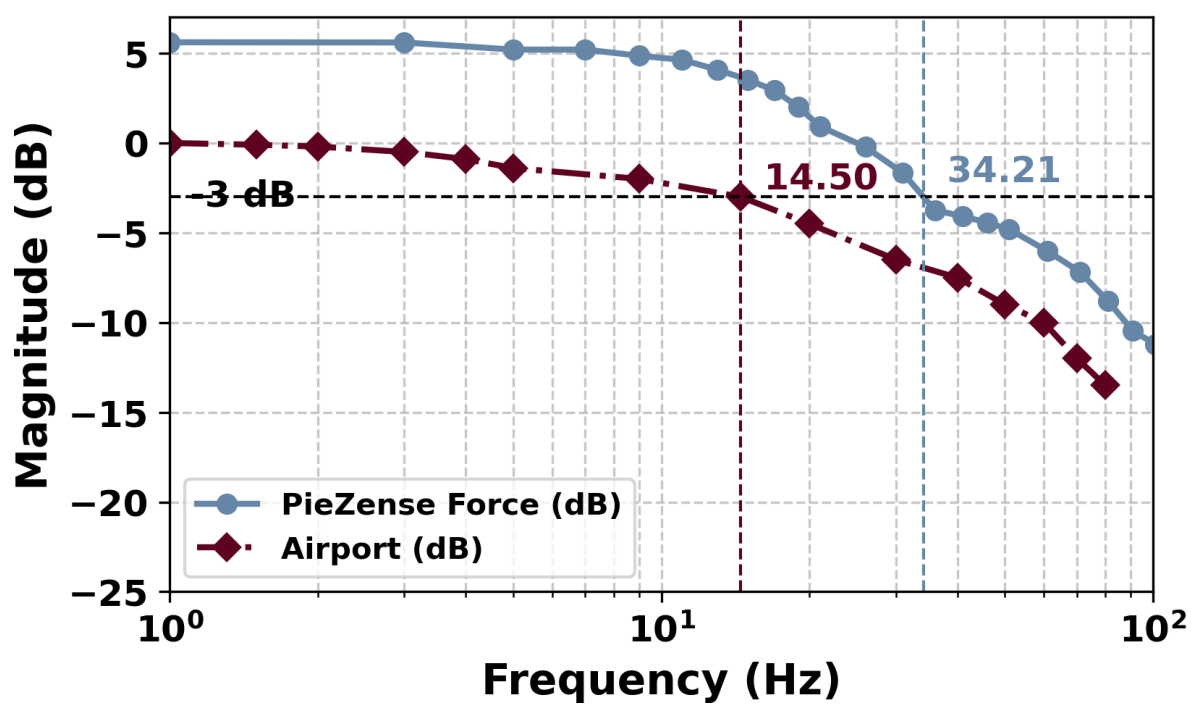


Figure S2: Force-frequency response of the closed-loop piezo-based device compared to a FlowIO controlled open-loop solenoid system used by du Pasquier et al. [2].

#### 1.3.2 Vertical displacement under load

To characterize how the textile actuator interacts with loaded surfaces, an experiment was conducted to measure its deformation under varying pressures and applied loads. The relationship between force and displacement provides insight into how the textile actuators interface with human skin.

**Methods** A vertical test stand similar to the one used in the force measurement tests is illustrated in Fig. S1c. Instead of a flat Plexiglas plate, a transparent Plexiglas tube was used to constrain the vertical motion of the load and ensure proper alignment with the actuator. A digital sliding caliper was mounted above the weight to measure vertical displacement with high precision. As the actuator was gradually pressurized to various pressure levels. As it inflated and lifted the weight upward, the displacement readings were recorded in real time. For each load condition, n = 2 tests were conducted to ensure consistency. The measurements were performed with n = 3 actuators ($n = 6$ tests totals) to account for actuator variability.

**Results** As shown in Fig. S3, actuator displacement increases non-linearly with applied pressure and is inversely proportional to the external load. The displacement generally decreases as the load increases. At lower loads, the displacement curve saturates more quickly, resulting in greater overall displacement and more pronounced nonlinearity. Conversely, higher loads tend to flatten and linearize the pressure–displacement response. Crucially, even under the highest tested load, the actuator achieves displacements exceeding 5 mm, enough to generate perceptible haptic feedback.

#### 1.3.3 Airtightness and pressure retention

Maintaining airtightness is essential for pneumatic systems to ensure efficient operation, minimize air and energy consumption, and enable reliable actuator control. To validate the airtightness of the platform, we tested the device's air retention systematically.

**Methods** In this experiment, three textile actuators were connected to a single device. Airtightness was then measured as follows:

1. The inlet port of the proportional valve was fully opened to allow unrestricted airflow from the $CO_2$ cartridge regulator to the actuators.

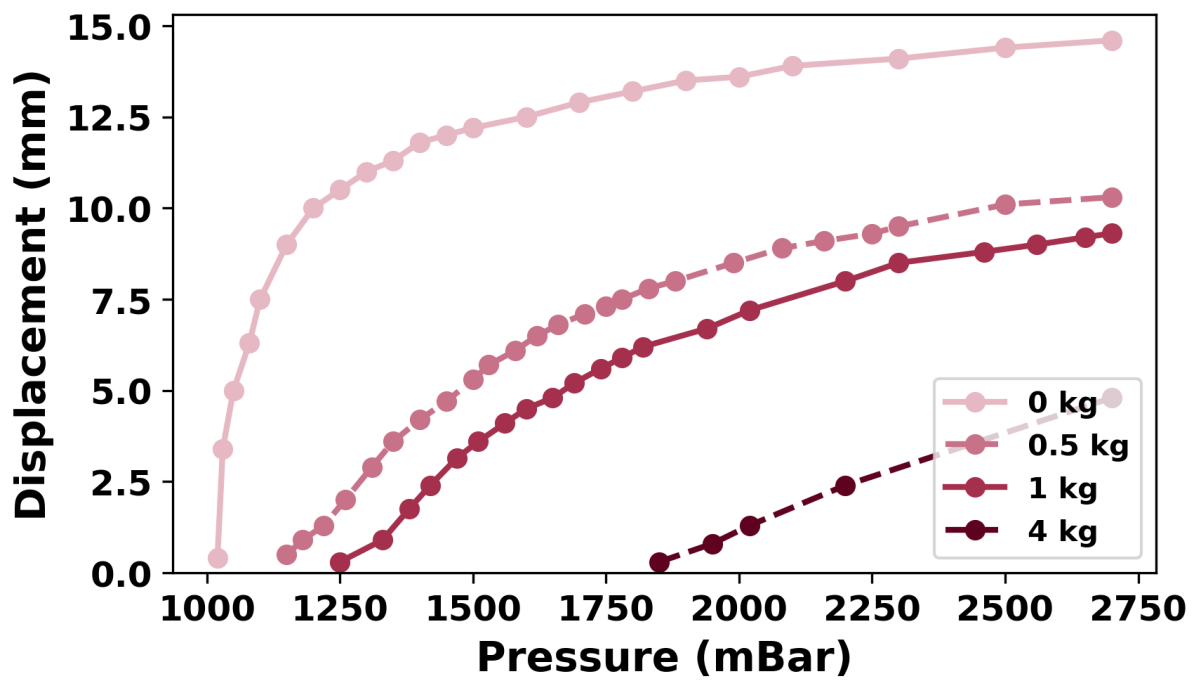


Figure S3: Average actuator displacement as a function of pressure under varying loads

2. A pressure regulator was adjusted to supply the standard operating pressure of 2665 mbar, fully inflating all actuators.
3. Once pressurized, the regulator was closed to isolate the device from the $CO_2$ source and prevent additional air intake.
4. The pressure gauge on the regulator was used to monitor internal pressure over time. The pressure decay was recorded to quantify airtightness.

The test was conducted twice ($n = 2$) to verify the repeatability and consistency of the results.

**Results** The device retained pressure with minimal leaks: it drops to 2300 mbar after 5 minutes, 1700 mbar after 16 minutes, and fully depressurizes only after 30 minutes. In contrast, a similar open-loop based on FlowIO [4] was tested using the same procedure and lost pressure in under 10 seconds.

### 1.3.4 Step response characterization

This experiment was used to characterize the transient behavior of the device by measuring its response time to a step input, which provided insight into its dynamic performance and control response.

**Methods** The device was initially stabilized at a baseline pressure of 1050 mbar. Then, a step pressure increase of 2000 mbar was applied. The resulting actuator response was recorded using the device's integrated pressure sensor, with time-resolved data collected via the GUI. This data was then analyzed using a custom Python script to extract key performance metrics such as rise time, settle time, and overshoot. To evaluate performance and repeatability, a total of 20 step response ($n = 40$) trials are conducted using two different valve–actuator pairs.

**Results** The resulting actuator response is shown in Fig. S4. The system exhibits an average rise time (10%–90%) of 29 ms (ranging from 5 ms to 34 ms), with a total settling time of 39 ms.
These results indicate that the controller achieved rapid and stable pressure control despite the inherent limitations of pneumatic systems, which are typically slower than vibrotactile actuators due to the compressibility of air and mechanical valve delays [5].

For wearable haptics, response times below 50 ms are generally sufficient [6] to render clear and timely tactile feedback. Thus, the device's responsiveness is well within the acceptable range for interactive applications and supports a variety of dynamic haptic effects such as buzzing, pulsing, tapping and squeezing.

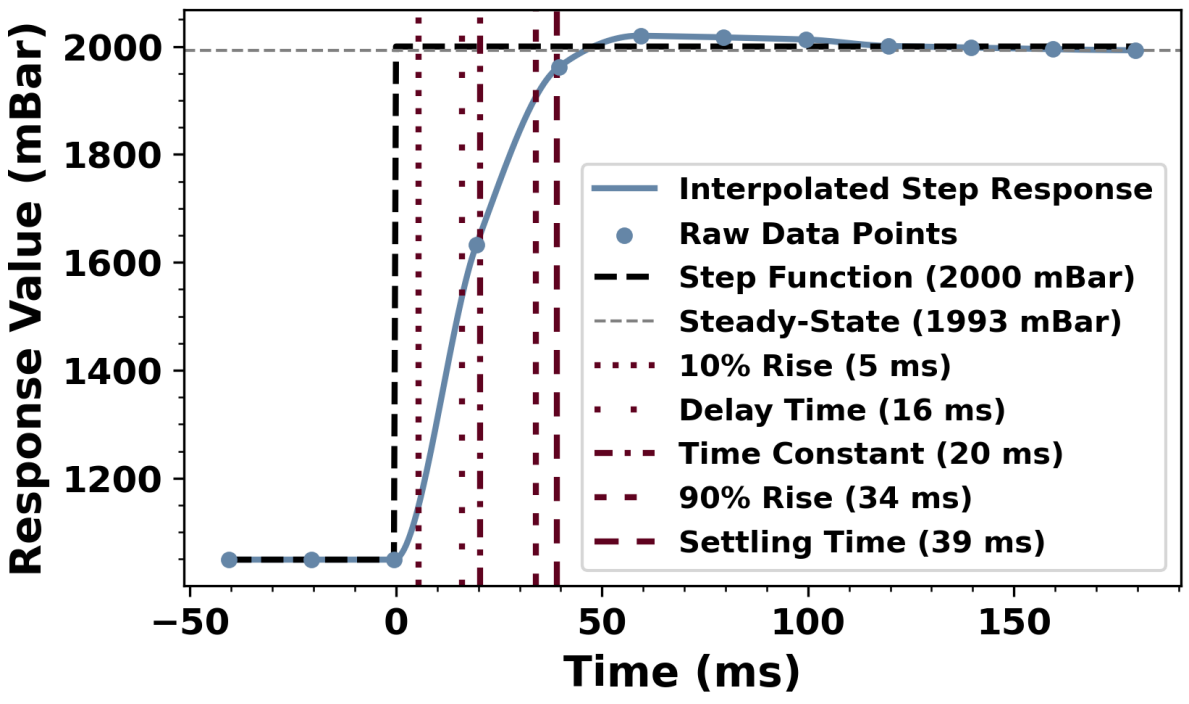


Figure S4: Pressure Step Response

## 1.4 Twin-system tactile communication evaluation

### 1.4.1 Wireless delay evaluation protocol

**Test setup and procedure** One device operated in sensing-only mode, monitoring pressure input from a pre-inflated actuator (1100 mbar) without actuation. The second device operated in listening mode and mirrored the sensed pressure data received wirelessly. A clear event was generated by rapidly compressing the sensing actuator in a single motion lasting approximately 80 ms, producing a sharp pressure spike. Pressure data from both devices were logged to CSV via the host GUI, with a shared wall-clock timestamp appended to each BLE packet, providing a common time reference across devices. The test was repeated ten times ($n = 10$). The process was additionally captured using an iPhone at 120 FPS to independently verify timing.

**Event onset detection and latency computation** Latency was computed from the synchronized CSV logs using a 3% step threshold method, consistent with the latency definition used in step response characterization (Section 1.3.4). Event onset on the sensing device was defined as the time at which the logged pressure trace first exceeded 3% of the peak pressure above the baseline sensing pressure of 1100 mbar, marking the onset of actuator compression. Event onset on the listening device was defined as the time at which the reproduced pressure trace first exceeded the same 3% threshold above its baseline. Latency per trial was computed as the difference between the two onset timestamps using the shared wall-clock reference. The reported mean latency of 64 ms represents the mean across $n = 10$ trials. The iPhone footage at 120 FPS was used to verify the plausibility of the computed values but was not used in the primary latency calculation.

### 1.4.2 Exploratory human communication task

**Participants and setup** Five participants (2 female, 3 male) took part in the bidirectional haptic communication task, forming user pairs for testing. In each session, one participant served as the sender (User S) and the other as the receiver (User R). The experiment was bidirectional: participants alternated roles so that both acted as sender and receiver during the session, requiring

no hardware changes between role switches. Each participant wore a device equipped with three textile-based actuators positioned along the forearm (labeled 1 to 3 from wrist to elbow). To eliminate visual bias, User R was blindfolded during all trials.

**Task conditions and trials** Three task conditions were tested. In the single actuator condition, one actuator was inflated to simulate a localized touch. In the simultaneous activation condition, two or more actuators were inflated together to simulate multi-point contact. In the intensity variation condition, a single actuator was inflated at varying pressure levels to simulate light, medium, and strong touch.

User R received 10 trials per condition ($n = 10$), for a total of 30 trials per participant pair. In each trial, User R was asked to identify the location, number of active actuators, or stimulus intensity depending on condition.

**Signal amplification** To improve perceptibility, transmitted signals were amplified by a fixed gain of $4\times$ at the GUI forwarding stage. For example, a pressure input increasing from 1100 mbar to 1200 mbar on the sender side produced a corresponding increase from 1100 mbar to 1500 mbar on the receiver's actuator.

**Results** User R identified the location and number of active actuators with 100% accuracy across all trials and conditions. Dynamic haptic patterns such as repeated pulses and oscillations were reliably perceived and correctly interpreted. Intensity classification across three levels achieved approximately 90% accuracy; the majority of errors occurred between the light and medium categories, which were occasionally perceived as similar in strength.

## 2 Review of Pneumatic Controllers for Wearables and Soft Robotics

Pneumatic actuation remains one of the dominant approaches for wearable robots, soft manipulators, and haptic interfaces due to its inherent compliance and high power density. However, the performance of such systems is fundamentally shaped by the design of the pneumatic controller, including its size, control strategy, sensing capabilities, and achievable dynamic response. Table S2 summarizes representative pneumatic control systems reported over the last 15 years for wearables, soft robotics, and haptics.

Table S2: **Comparison of pneumatic control platforms and representative wearable and soft robotics systems.** NR (not reported) indicates the source does not specify the metric.

| Source | Weight | Pressure source | Valve Type | Channels | Pressure range | Closed loop? | Performance metrics | Applications |
|---|---|---|---|---|---|---|---|---|
| **Controller platforms** | | | | | | | | |
| FlowIO [7, 2] | 370 g | Air/vacuum pump | Solenoid | Up to 8 | -180 kPa to 200 kPa | No | 14.5 Hz<br>3.2 L/min | Modular, portable controller. |
| PneuDuino [8] | NR | Wall air | Solenoid | 1 | 400 kPa | No | NR | Open source pneumatic platform |
| Programmable Air [9] | NR | Air/vacuum pump | Solenoid | 1 | -50 kPa to 50 kPa | No | 2 L/min | Open source pneumatic platform |
| Soft Robotics Toolkit [10] | NR | Air pump | Solenoid | 4 to 8 | 100 kPa | Possible | NR | Soft robotics laboratory control |
| Syringe Control System[11] | NR | Syringe + linear motor | None | up to 8 | 175 kPa | Possible | NR | Low-cost pneumatic control |
| **Application-driven systems** | | | | | | | | |
| Polygerinos et al [12] | 3.3 kg | Hydraulic pump (offboard) | Solenoid | 8 | 400 kPa | No | 0.25 Hz | Wearable actuation system for rehabilitation |
| Cappello et al [13] | NR | Wall air | Proportional valves | Multiple | NR | NR | NR | Soft exosuit for assistance for mobility support |
| Qi et al [14] | 283 g | Air pump + reservoir | Solenoid | 5 | 10 to 100 kPa | Yes | NR | Haptic glove |
| He et al. [15] | NR | Air pump | Solenoid | 5 | NR | No | NR | Haptic bracelet |
| Jumet et al. [16] | 345 g | Air cartridge | Solenoid | 1 to 3 | up to 100 kPa | No | 10-20 Hz | Haptic sleeve |

Across these systems, most wearable and soft robotic platforms rely on either external pneumatic supplies or compact micropumps, with closed-loop pressure control increasingly adopted to improve stability and repeatability. However, reported systems typically focus on unidirectional actuation and do not address sustained, bidirectional tactile signal exchange between remote locations, particularly in untethered and wearable form factors.

# 3 Teleoperation Control Implementation

This section provides full implementation details for the teleoperation system used during the user studies.

## 3.1 Hand and Wrist Pose Tracking

Human hand and wrist poses were tracked using a Microsoft HoloLens 2 mixed-reality headset. The headset provides six-degree-of-freedom tracking and real-time hand pose estimation derived from a combination of a time-of-flight depth sensor, two infrared cameras, and an onboard inertial measurement unit. These sensing modalities enable robust estimation of wrist position and orientation, as well as fingertip locations, without the need for external cameras or additional calibration.

The see-through optical display allows users to view the physical environment directly, avoiding reliance on video passthrough, which has been shown to contribute to discomfort and fatigue in immersive teleoperation systems. The head-mounted form factor also allows users to move freely within the workspace while maintaining consistent tracking performance.

## 3.2 Arm Control Mapping

The robotic platform used in this study was a Kinova Gen3 manipulator controlled through Kinova's ROS driver. Cartesian velocity commands were sent to the robot end-effector, and joint velocities were computed internally using Kinova's proprietary inverse kinematics solver.

Upon launching the teleoperation application on the HoloLens 2, the device's world coordinate frame is initialized. To approximately align the device world frame with the robot base coordinate frame, the application is started while standing directly behind the robot and looking forward. Because control commands are based on pose deltas rather than absolute positions, precise collocation of the coordinate frames is not required. Empirical testing indicated that this alignment convention is sufficient for stable control. Equations S6 and S7 describe the velocity relationships between the two stationary frames for the default control mode and when participants view the workspace through the wrist-mounted camera feed.

To map human wrist motion to robot motion, linear velocities were computed from the change in wrist position $(x_{\text{wrist}}, y_{\text{wrist}}, z_{\text{wrist}})$ expressed in a stationary HoloLens world frame and scaled by the elapsed time between detections. Angular velocities were computed by taking the difference between consecutive wrist orientations represented as quaternions, converting this difference to a rotation vector, and scaling it by the same time interval. The calculations are detailed in Equations S1–S8.

During all teleoperation tasks, the gripper orientation was constrained to a top-down grasp configuration. As a result, only angular velocities about the gripper's $Z$ axis were permitted. This constraint reduced task complexity and ensured consistent contact geometry during manipulation.

$$\Delta t = t_k - t_{k-1} \tag{S1}$$

$$\Delta \mathbf{p}^{\text{Holo}} = \begin{bmatrix} \Delta x \\ \Delta y \\ \Delta z \end{bmatrix} = \begin{bmatrix} x_k - x_{k-1} \\ y_k - y_{k-1} \\ z_k - z_{k-1} \end{bmatrix} \tag{S2}$$

$$\mathbf{v}^{\text{Holo}} = \frac{\Delta \mathbf{p}^{\text{Holo}}}{\Delta t} = \frac{1}{\Delta t} \begin{bmatrix} x_k - x_{k-1} \\ y_k - y_{k-1} \\ z_k - z_{k-1} \end{bmatrix} \tag{S3}$$

$$\mathbf{q}_\Delta = \mathbf{q}_k \mathbf{q}_{k-1}^{-1} \tag{S4}$$

$$\boldsymbol{\omega}^{\text{Holo}} = \frac{\text{rotvec}(\mathbf{q}_\Delta)}{\Delta t} \tag{S5}$$

$$\mathbf{v}_{\text{cmd}}^{\text{robot}} = \begin{cases} \begin{bmatrix} v_z^{\text{Holo}} \\ -v_y^{\text{Holo}} \\ v_x^{\text{Holo}} \end{bmatrix}, & \text{camera-feed mode} \\ \begin{bmatrix} v_x^{\text{Holo}} \\ v_y^{\text{Holo}} \\ v_z^{\text{Holo}} \end{bmatrix} & \text{default} \end{cases} \tag{S6}$$

$$\boldsymbol{\omega}_{\text{cmd}}^{\text{robot}} = \begin{bmatrix} 0 \\ 0 \\ -\omega_z^{\text{Holo}} \end{bmatrix} \tag{S7}$$

(S8)

### 3.3 Gripper Control and Hand Calibration

Control of the parallel-jaw gripper was achieved by mapping the distance between the user's thumb and index finger to the gripper opening width. Because hand size varies across users, an initial calibration step was performed at the beginning of each session.

During calibration, participants were instructed to open their thumb and index finger to a maximally comfortable distance, denoted $d_{\text{fingers,max}}$. This value was mapped to the fully open state of the gripper. During teleoperation, the instantaneous distance between the detected thumb and index finger positions, $d_{\text{fingers}}$, was continuously computed. The normalized ratio was then used to generate the gripper command:

$$g_{\text{cmd}} = 1 - \frac{d_{\text{fingers}}}{d_{\text{fingers,max}}},$$

where $g_{\text{cmd}} = 0$ corresponds to a fully open gripper and $g_{\text{cmd}} = 1$ corresponds to a fully closed gripper.

To ensure smooth and stable gripper motion, gripper commands were updated only when three conditions were simultaneously satisfied: (1) the absolute change from the previous command exceeded 0.05, (2) at least 0.01 s had elapsed since the last update, and (3) the translational velocity of the robot end-effector was below 0.05 m/s. Condition (1) was determined empirically to prevent unintended gripper activation by filtering out natural, unintentional finger movements or oscillations during arm motion. Condition (2) was also determined empirically to filter noise and ensure

intentional commands at a consistent rate. Finally, condition (3) was defined a priori to restrict finger-based commands to moments when the arm was near rest, thereby reducing false gripper activations caused by finger-tracking errors while the participant's hand or the robotic arm was in motion.

# 4 Detailed User Study Procedures and Additional Results

## 4.1 Participant Instructions and Task Flow

$N = 25$ participants were recruited through university email lists and Slack channels. Participants provided informed consent, and the protocol received approval from the Stanford University Institutional Review board (#65022).

Participants completed the study in two phases: Perception and Teleoperation. Prior to beginning each phase, participants received standardized verbal instructions explaining the task objectives and response format. Participants were encouraged to ask clarification questions before each task began.

During perception tasks, participants were seated facing away from the robot to eliminate visual cues. During teleoperation tasks, participants stood in designated positions corresponding to each visual condition.

## 4.2 Perception Task Parameters

### 4.2.1 Force Discrimination

Three discrete pressure levels (150, 300, and 450 mbar) were delivered to the fingertip actuators in randomized order. Participants performed both 2-element and 3-element comparison tasks, reporting whether the second stimulus was greater than, smaller than, or equal to the reference.

### 4.2.2 Stiffness Discrimination

Three visually identical lattice cubes of increasing stiffness were fabricated using a Formlabs Form 3 printer and Elastic resin. Stiffness was measured in compression using an Instron 68TM-50 equipped with a 5 kN static load cell. The cubes were compressed to 10% of their original height at a rate of 0.6 mm/min. The measured stiffness values were 0.630 N/mm, 1.77 N/mm, and 3.35 N/mm. Participants completed both direct manual discrimination and robot-mediated discrimination under fixed gripper closure conditions.

### 4.2.3 Weight Discrimination

Three visually identical weights (150 g, 300 g, and 450 g) were used. In the teleoperated condition, participants controlled both the robot arm and gripper while attempting to grasp and lift each object with minimal applied force.

## 4.3 Teleoperation Task Parameters

Participants completed three pick-and-place tasks under different visual conditions. For PnP1, participants had an unobstructed view of the robot and workspace. For PnP2, visual occlusions were introduced due to robot self-occlusion. For PnP3, participants relied exclusively on the wrist-mounted camera feed displayed on an external monitor.

Table S3: Path Lengths (NH vs H) across tasks. Values are reported as Mean ± SD. Differences are computed as Δ (NH–H). Paired t-tests were performed, and effect sizes are reported as Cohen's $d_z$.

| **Task** | **NH** | **H** | **Δ(NH–H)** | $t$ | $p$ | $d_z$ |
|---|---|---|---|---|---|---|
| PnP1 | $0.82 \pm 0.39$ | $0.70 \pm 0.27$ | 0.12 | 1.57 | 0.133 | 0.36 |
| PnP2 | $0.78 \pm 0.35$ | $0.77 \pm 0.28$ | 0.02 | 0.26 | 0.799 | 0.06 |
| PnP3 | $1.52 \pm 1.03$ | $1.08 \pm 0.37$ | 0.44 | 1.84 | 0.082 | 0.42 |

Table S4: Trial Durations (NH vs H) across tasks. Values are reported as Mean ± SD. Differences are computed as Δ (NH–H). Paired t-tests were performed, and effect sizes are reported as Cohen's $d_z$.

| **Task** | **NH** | **H** | **Δ(NH–H)** | $t$ | $p$ | $d_z$ |
|---|---|---|---|---|---|---|
| PnP1 | $12.88 \pm 6.86$ | $10.08 \pm 4.51$ | 2.81 | 1.80 | 0.088 | 0.41 |
| PnP2 | $10.99 \pm 7.17$ | $11.11 \pm 6.14$ | −0.12 | −0.09 | 0.933 | −0.02 |
| PnP3 | $22.74 \pm 16.45$ | $16.91 \pm 6.78$ | 5.83 | 1.41 | 0.175 | 0.32 |

Each participant completed n = 10 trials per condition for PnP1 and n = 5 trials per condition for PnP2 and PnP3. Condition order was randomized within each task.

## 4.4 Metrics and Data Processing

**Accuracy** During perception tasks, true and predicted stimulus sequences were recorded and compared using confusion matrices. Accuracy was computed as the percentage of correctly identified relationships across all trials.

**Task Completion Time** Completion time was defined as the interval between initial contact with the object at the start location and release of the object at the goal location. Trials exceeding 5 minutes were marked as failures.

**Path Length** Gripper position $(x, y, z)$ was logged continuously, and total path length was computed as the sum of Euclidean distances between consecutive samples.

**Perceived Usability** System usability was assessed using the System Usability Scale (SUS).

**Perceived Task Load** Perceived workload was assessed using the NASA Task Load Index (TLX), including all six subscales.

## 4.5 Statistical Analysis

All statistical analyses were conducted in Python using the SciPy statistics package. Data normality was assessed using the Shapiro–Wilk test. Following prior work [17, 18], paired t-tests were applied to compare conditions with and without haptic feedback, regardless of normality. Trials with missing data resulting from logging errors were excluded symmetrically across conditions.

Table S5: Paired t-test results comparing the NH and H conditions across NASA Task Load Index (TLX) workload subscales ($n = 19$). Values are reported as mean ± standard deviation. Shapiro–Wilk tests were conducted on paired difference scores to assess normality assumptions. Significant effects are highlighted in bold.

| **Measure** | **NH** | **H** | **Δ(NH − H)** | $t$ | $p$ | $d_z$ |
|---|---|---|---|---|---|---|
| Mental Demand | 47.63 ± 28.98 | 38.95 ± 23.13 | 8.68 | 1.96 | 0.065 | 0.45 |
| Physical Demand | 43.68 ± 25.43 | 43.16 ± 24.79 | 0.53 | 0.26 | 0.797 | 0.06 |
| Temporal Demand | 33.95 ± 27.06 | 30.53 ± 21.66 | 3.42 | 1.36 | 0.190 | 0.31 |
| Performance | 46.84 ± 21.74 | 46.84 ± 25.07 | 0.00 | 0.00 | 1.000 | 0.00 |
| Effort | 54.21 ± 24.05 | 52.11 ± 24.51 | 2.11 | 0.82 | 0.424 | 0.19 |
| **Frustration** | **52.89 ± 22.63** | **45.26 ± 21.70** | **7.63** | **2.81** | **0.012** | **0.65** |

Normality of paired difference scores was assessed using the Shapiro–Wilk test. The normality assumption was violated for Temporal Demand ($p < .001$), Performance ($p = .007$), and Effort ($p = .002$), while the remaining measures satisfied normality assumptions.

Table S6: Overall System Usability Scale (SUS) scores comparing the NH and H conditions ($n = 19$). Values are reported as mean ± standard deviation.

| **Measure** | **NH** | **H** | **Δ(NH − H)** | $t$ | $p$ | $d_z$ |
|---|---|---|---|---|---|---|
| SUS | 61.05 ± 14.61 | 64.87 ± 16.32 | −3.82 | 1.79 | 0.090 | −0.41 |

### 4.6 Additional Results

**Per participant Teleoperation results** Here, we report per-participant mean applied pressure, task duration, and gripper path length for each of the three teleoperation tasks. Participants were randomly assigned to begin each task either with or without haptic feedback. These results are shown in Figs. S5, S6, and S7 for the Direct View (PnP1), Obstructed View (PnP2), and Camera-Only View (PnP3) conditions, respectively.

As reported in the main text, the addition of haptic feedback led to overall improvements across all three performance metrics when averaged across participants. Examination of the per-participant results reveals additional structure. In PnP1 and PnP2, the order in which participants experienced the conditions influenced task duration and path length: participants tended to take longer and follow longer trajectories in the condition they encountered first, regardless of whether haptic feedback was present. This ordering effect was not observed in PnP3 (Camera-Only View), where visual uncertainty dominated performance.

Across all tasks and conditions, the applied pressure was consistently lower or comparable when haptic feedback was provided, indicating that haptic cues supported force regulation independently of task order effects.

**Time allocation results between Pick vs. Place phases during teleoperation** Here, we report the average time allocations between Pick and Place phases during the pick-and-place tasks. The Pick phase was defined as approaching and executing a grasp around the object. The Place phase was defined as transporting the grasped object to the target location. The transition point between Pick and Place phases was defined by the onset of sustained motion after grasping the

Table S7: Time allocations (Pick vs Place) across tasks and haptic condition (NH vs H). Log ratios $log(T_{pick}/T_{place})$ are computed per trial then aggregated across participants. Values are reported as Mean ± SD. Differences are computed as Δ (NH–H). Paired t-tests were performed, and effect sizes are reported as Cohen's $d_z$. Normality assumptions of paired difference scores are assessed by Shapiro Wilk tests.

| **Task** | **NH** | **H** | **Δ(NH–H)** | $t$ | $p$ | $d_z$ |
|---|---|---|---|---|---|---|
| PnP1 | $+0.021 \pm 0.702$ | $+0.040 \pm 0.714$ | $-0.018$ | $-0.12$ | 0.907 | $-0.03$ |
| PnP2 | $+0.007 \pm 0.873$ | $-0.081 \pm 0.797$ | $+0.088$ | $+0.72$ | 0.481 | $+0.17$ |
| PnP3 | $+0.400 \pm 0.816$ | $+0.036 \pm 0.856$ | $+0.365$ | $+2.09$ | 0.052 | $+0.49$ |
| All | $+0.087 \pm 0.685$ | $+0.003 \pm 0.631$ | $+0.084$ | $+1.08$ | 0.296 | $+0.25$ |

object. The log ratio of Pick vs. Place durations, $log(T_{place}/T_{pick})$ was computed for each trial; symmetric trials under no-haptics (NH) and haptics (H) conditions were compared to express the haptic effect as a within-participant paired difference per task (PnP1, PnP2, and PnP3) conditions, which were then aggregated across participants. Absolute within-trial phase durations were also computed. These results are shown in Fig. S8.

Across all tasks and conditions, presence of haptic feedback consistently shifted time allocation toward the Pick phase. Participants spent proportionally less of each trial to transporting the object (and correspondingly more relative time on grasping) when haptic feedback was available. The time shifts are primarily due to a faster transport rather than a slower grasp, indicating that the task efficiency improvements discussed in the main text can be attributed to participants moving the object more decisively after committing to a grasp.

**SUS Results** Here, we report the average SUS scores for each condition (Figure S9). The SUS is a standardized 10-item Likert-scale questionnaire used to evaluate perceived system usability. Odd-numbered items are positively framed, for which higher scores indicate better usability, whereas even-numbered items are negatively framed, for which lower scores indicate better usability.

As discussed in the main text, participants perceived the system with haptic feedback as more usable overall, although the difference did not reach statistical significance. This trend nevertheless suggests that haptic feedback may positively influence users' perceptions of system usability and interaction quality. The absence of statistical significance is likely attributable, in part, to the additional complexity introduced by integrating the haptic feedback subsystem with the robotic platform, which may have increased the overall cognitive and operational demands of the system. Future work will focus on streamlining the system architecture and improving integration robustness to better realize the usability benefits of haptic feedback.

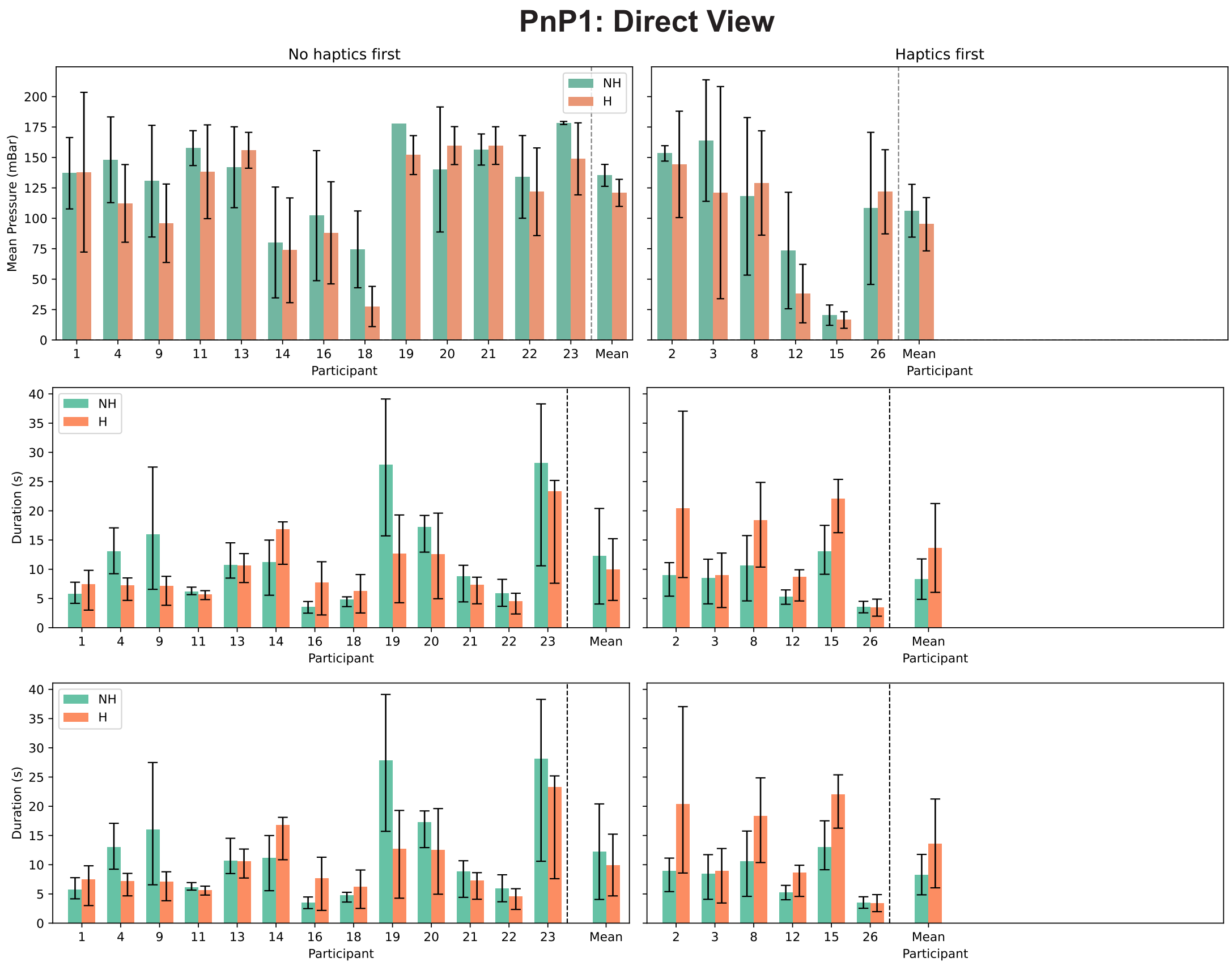


Figure S5: **Per Participant Results for the Direct View Teleoperation Task** Mean applied pressure, task duration, and gripper path length are reported for each participant under no-haptics (NH) and haptics (H) conditions; participants were randomly assigned to begin the experiment either with or without haptic feedback and are separated into left and right panels based on their starting condition.

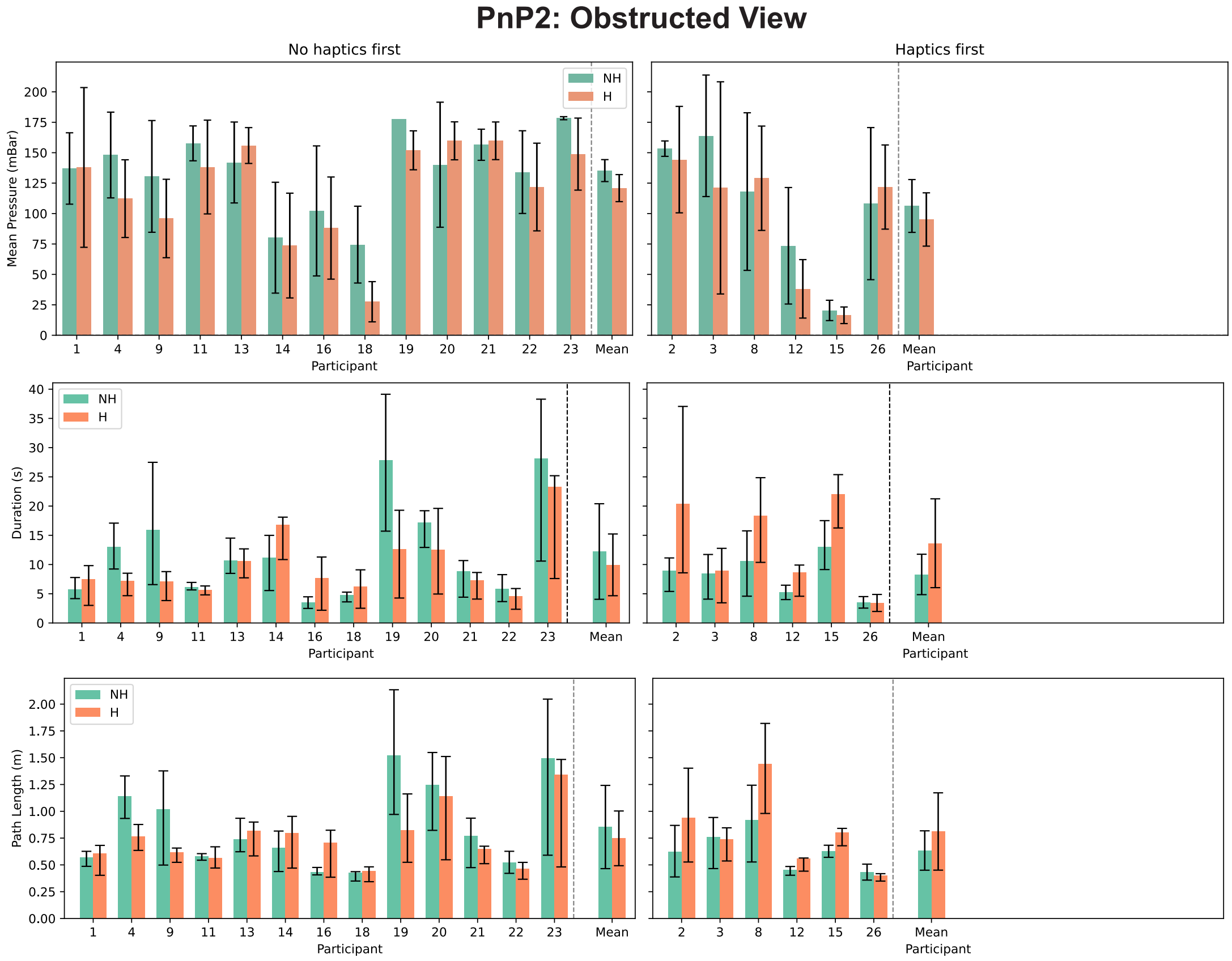


Figure S6: **Per Participant Results for the Obstructed View Teleoperation Task** Mean applied pressure, task duration, and gripper path length are reported for each participant under no-haptics (NH) and haptics (H) conditions; participants were randomly assigned to begin the experiment either with or without haptic feedback and are separated into left and right panels based on their starting condition.

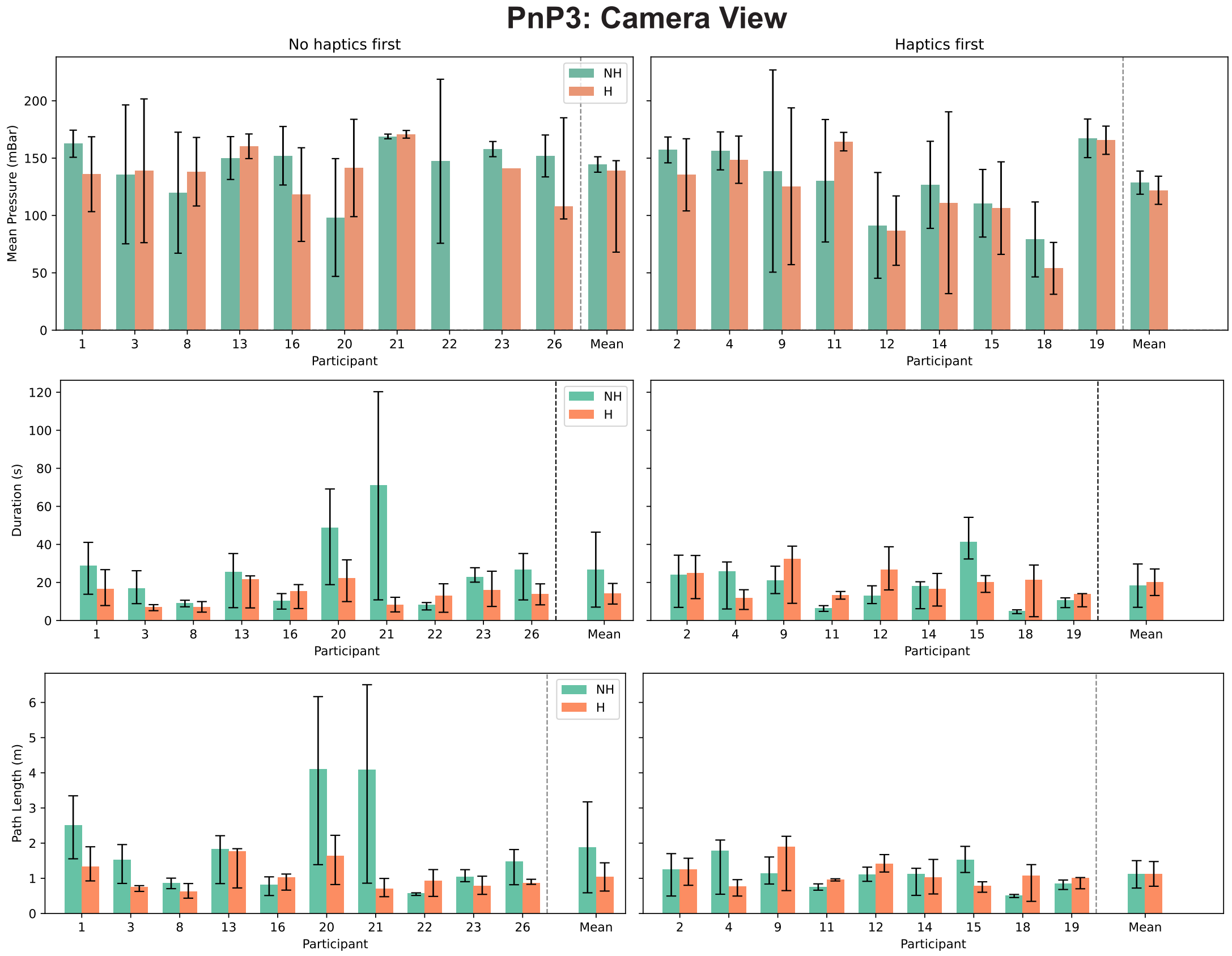


Figure S7: **Per Participant Results for the Camera View Teleoperation Task** Mean applied pressure, task duration, and gripper path length are reported for each participant under no-haptics (NH) and haptics (H) conditions; participants were randomly assigned to begin the experiment either with or without haptic feedback and are separated into left and right panels based on their starting condition.

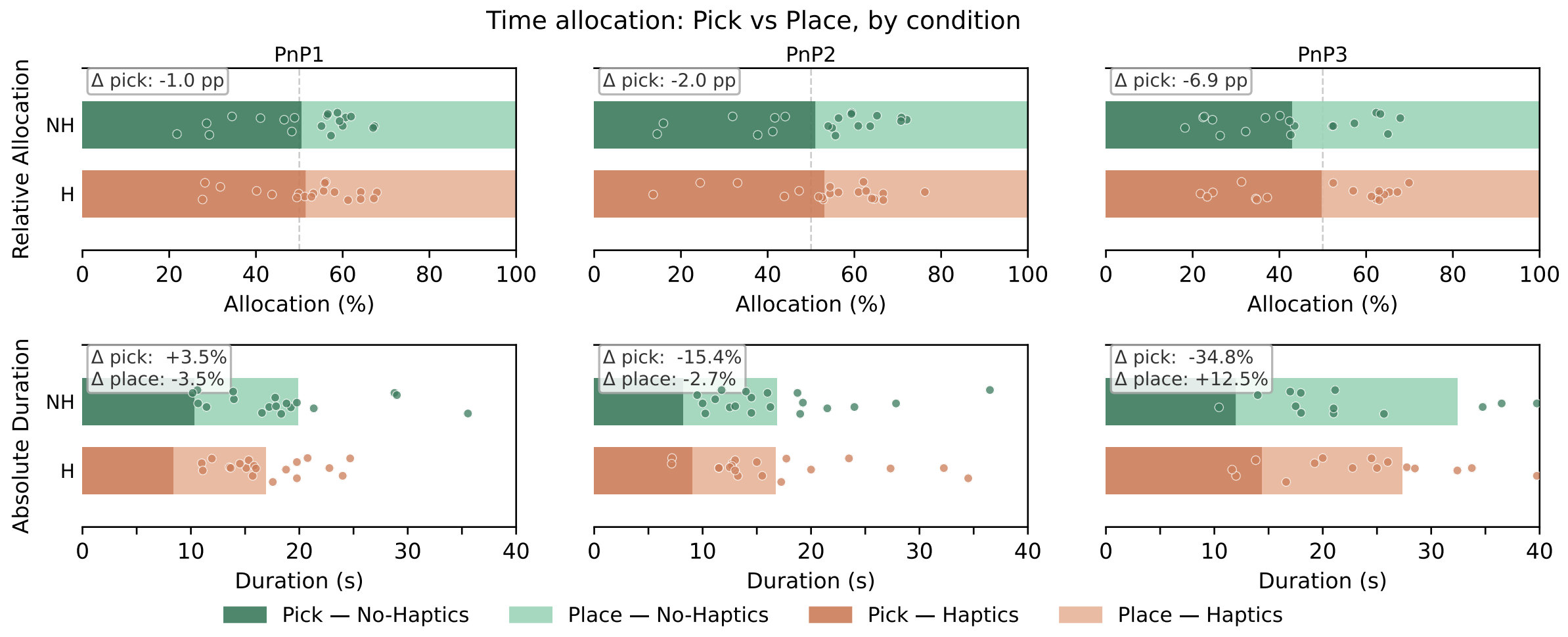


Figure S8: **Pick vs. Place time allocation by condition.** Upper row: trial movement time spent in Pick and Place phases (excludes inter-trial return transit) per condition. Vertical line marks equal (50%) allocation. Inset annotation reports absolute NH–H percentage point shift between no-haptics (NH) and haptics (H) conditions. Lower row: absolute stacked durations of trial movement time. Bars show group means of participant means; dots are individual participants. Inset annotation reports percent change from NH baseline. All Δ annotations are mean paired within-participant shifts (change under haptics).

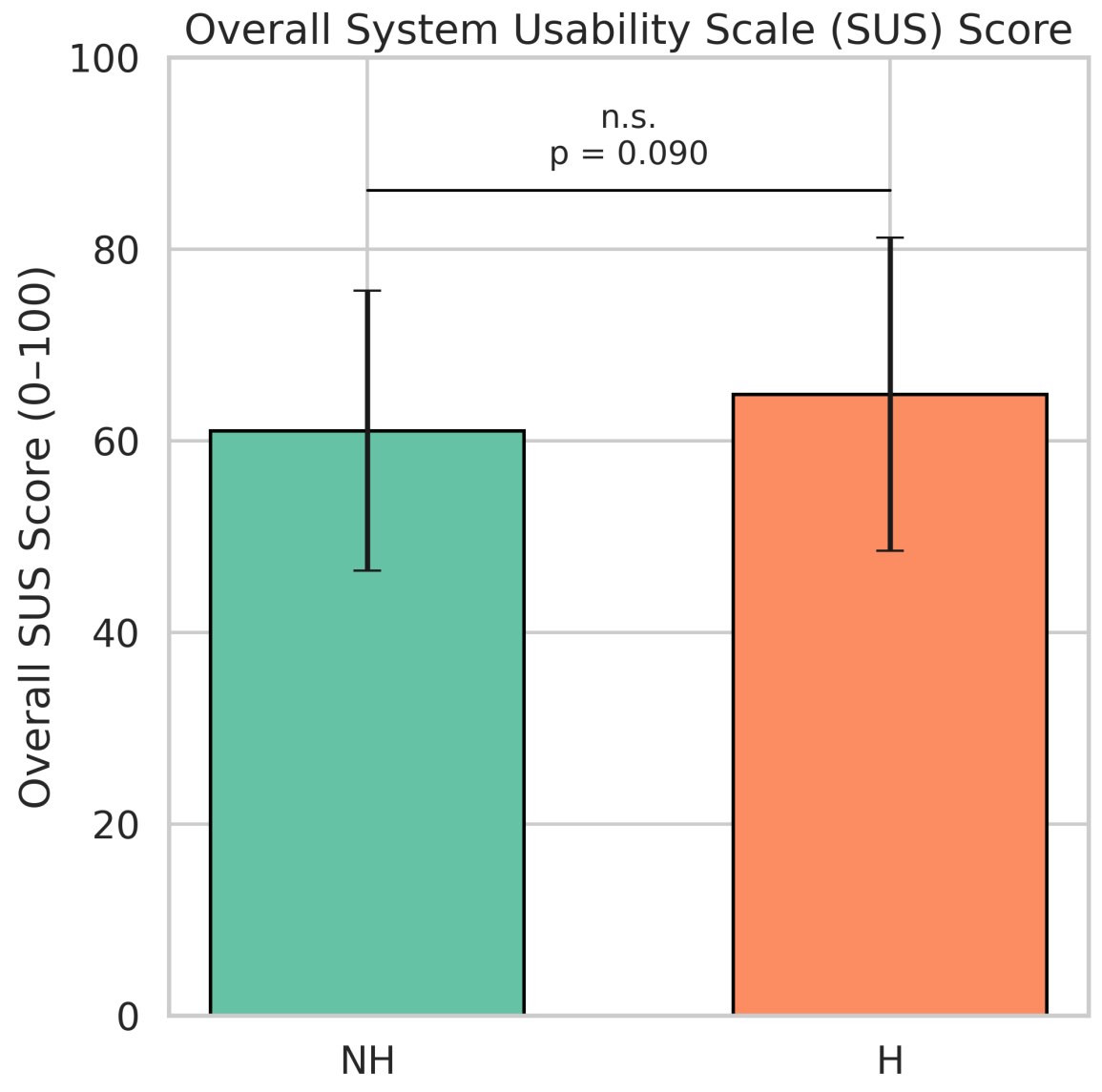


Figure S9: **SUS**: Average scores for each condition (n = 19).





\newpage

\end{document}